\pdfoutput=1
\documentclass[11pt]{article}

\usepackage{acl}

\usepackage{times}
\usepackage{latexsym}

\usepackage[T1]{fontenc}
\usepackage[utf8]{inputenc}
\usepackage[russian,english]{babel}
\usepackage{subcaption}
\usepackage{microtype}
\usepackage{booktabs} 
\usepackage{multirow}
\usepackage{hyperref}
\usepackage{xcolor}
\usepackage{url}
\usepackage{caption}
\usepackage{tabularx}
\usepackage{multicol}
\usepackage{floatrow}
\usepackage{microtype}
\usepackage{amsmath}
\usepackage{xspace}
\usepackage{amssymb}
\usepackage{adjustbox}
\usepackage{textcomp}
\usepackage{colortbl}
\usepackage{arydshln}
\usepackage{wrapfig}
\usepackage{todonotes}
\usepackage{arydshln}
\usepackage{euflag}
\makeatletter
\newcommand*\iftodonotes{\if@todonotes@disabled\expandafter\@secondoftwo\else\expandafter\@firstoftwo\fi}
\makeatother

\definecolor{grey}{gray}{0.92}

\newcommand{\datasetacronym}{\textsc{cod}\xspace}
\newcommand{\cod}{\textsc{cod}\xspace}
\newcommand{\tod}{\textsc{ToD}\xspace}

\newcommand{\rparagraph}[1]{\vspace{1.2mm}\noindent\textbf{#1.}}

\definecolor{Gray}{gray}{0.92}
\newcolumntype{Y}{>{\centering\arraybackslash}X}

\title{Cross-Lingual Dialogue Dataset Creation via Outline-Based Generation}

\author{Olga Majewska$^1$\thanks{~~Equal contribution.} ~~~ Evgeniia Razumovskaia$^{1}$\footnotemark[1] ~~~ Edoardo Maria Ponti$^{2,3}$ \\ {\bf ~~ Ivan Vuli\'{c}$^1$ ~~~ Anna Korhonen$^1$} \smallskip \\
$^1$Language Technology Lab, TAL, University of Cambridge \\
$^2$Mila -- Quebec Artificial Intelligence Institute~~~$^3$McGill University\\
$^1$\texttt {\{om304,er563,iv250,alk23\}@cam.ac.uk} \\
$^2$\texttt {edoardo-maria.ponti@mila.quebec}
}

\begin{document}
\maketitle
\begin{abstract}
\textit{Multilingual} task-oriented dialogue (\tod) facilitates access to services and information for many (communities of) speakers. Nevertheless, the potential of this technology is not fully realised, as current datasets for multilingual \tod---both for modular and end-to-end modelling---suffer from severe limitations. \textbf{1)} When created from scratch, they are usually small in scale  and fail to cover many possible dialogue flows. \textbf{2)} Translation-based \tod datasets might lack naturalness and cultural specificity in the target language. In this work, to tackle these limitations we propose a novel \textit{outline-based} annotation process for multilingual \tod datasets, where domain-specific abstract schemata of dialogue are mapped into natural language outlines. These in turn guide the target language annotators in writing a dialogue by providing instructions about each turn's intents and slots. Through this process we annotate a new large-scale dataset for training and evaluation of multilingual and cross-lingual \tod systems. Our \textbf{C}ross-lingual \textbf{O}utline-based \textbf{D}ialogue dataset (termed \datasetacronym) enables natural language understanding, dialogue state tracking, and end-to-end dialogue modelling and evaluation in 4 diverse languages: Arabic, Indonesian, Russian, and Kiswahili. Qualitative and quantitative analyses of \datasetacronym{} versus an equivalent translation-based dataset demonstrate improvements in data quality, unlocked by the outline-based approach. Finally, we benchmark a series of state-of-the-art systems for cross-lingual \tod, setting reference scores for future work and demonstrating that \datasetacronym prevents over-inflated performance, typically met with prior translation-based \tod datasets.

%%revealing that `translate-test' baselines are superior to multilingual models for both intent classification and end-to-end modelling, but not for slot labelling. In both cases, the performance on outline-based dialogues is more modest compared to translation-based dialogues, as the latter is arguably over-inflated.

%% (IV), keep it for the intro: (due to being time- and cost-intensive)
%% (IV), keep it for the intro: (due to the open-endedness of human-to-human interaction)
%% (IV) uman respondents to a survey favoured outline-based dialogues 8 times out of 10 and gave higher scores for both their naturalness and familiarity of their concepts / entities. 

\end{abstract}

\section{Introduction and Motivation}
% Multilingual NLU, DST, and e2e in dialogue
One of the staples of machine intelligence is arguably the ability to communicate with humans and complete a task as instructed during such an interaction. This is commonly referred to as task-oriented dialogue \citep[\tod;][]{gupta2005t,bohus2009ravenclaw,young2013pomdp,muise2019planning}. Despite having far-reaching applications, such as banking \citep{altinok2018ontology}, travel \citep{zang2020multiwoz}, and healthcare \citep{denecke2019intelligent}, this technology is currently limited to a handful of languages \citep{razumovskaia2021crossing}. Thus, large communities of speakers are prevented access to automated services and information.

% DATA PAUCITY AND TRANSFER LEARNING (merge with above?)
 The progress in multilingual \tod is critically hampered by the paucity of training data for many of the world's languages. While cross-lingual transfer learning \cite{zhang2019jointlearningwithbert,xu2020multiatis++,siddhant2020evaluating,krishnan2021multilingual} offers a partial remedy, its success is tenuous beyond typologically similar languages and generally hard to assess due to the lack of evaluation benchmarks \citep{razumovskaia2021crossing}. What is more, transfer learning often cannot leverage multi-source transfer and few-shot learning due to lack of language diversity in the \tod datasets \citep{zhu2020crosswoz,quan2020risawoz,farajian-etal-2020-findings}.
 
 \begin{table*}
\def\arraystretch{0.9}
{\footnotesize
\begin{tabularx}{1.0\textwidth}{p{0.55\textwidth}p{0.4\textwidth}}
\toprule
\rowcolor{Gray}
\textbf{Outlines} & \textbf{Dialogue \& Slot Output}\\ 
\cmidrule(lr){1-1} \cmidrule(lr){2-2}
\textbf{USER:} \textit{Express the desire to search for roundtrip flights for a trip} &
\foreignlanguage{russian}{Мне нужно найти рейс в Ставрополь и обратно авиакомпании S7.}\\
\hdashline
the name of the airport or city to arrive at: Seattle & \foreignlanguage{russian}{Ставрополь}\\
the company that provides air transport services: American Airlines & S7\\
\cmidrule(lr){1-1} \cmidrule(lr){2-2}

\textbf{ASSISTANT/SYSTEM:} \textit{Inform the user that you found 1 such option(s). Offer the following option(s):}	& \foreignlanguage{russian}{Найден 1 рейс авиакомпании S7 с пересадкой, вылет в 7:35, возвращение в Москву в 16:15. Стоимость билетов 6845 рублей.}\\
\hdashline
the company that provides air transport services:  American Airlines & S7\\
departure time of the flight flying to the destination:  7:35am	& 07:35\\
departure time of the flight coming back from the trip:  4:15pm	& 16:15\\
the total cost of the flight tickets:  \$343	& \foreignlanguage{russian}{6845 рублей}\\

\bottomrule
\end{tabularx}
}%
\vspace{-1mm}
\caption{Example from the \datasetacronym dataset of outline-based dialogue generation in Russian with target language substitutions of slot values. The first column (\textbf{Outline}) includes example outlines presented to the dialogue creators, and the second column holds the creators' output (\textbf{Dialogue \& Slot Output}).}
\label{tab:prompt_examples}
\vspace{-2.5mm}
\end{table*}
 
 %% To exacerbate this situation

% LIMITATIONS OF TRANSLATION
Therefore, the main driver of development in multilingual \tod is the creation of multilingual resources. However, even when available, these resources suffer from several pitfalls. Most are obtained by manual or semi-automatic translation of an English source \citep[\textit{inter alia}]{castellucci2019slu-it,bellomaria2019almawave,susanto2017neural,upadhyay2018multiatis,xu2020multiatis++,ding2021globalwoz,Zuo:2021allwoz}. While this process is cost-efficient and typically makes data and results comparable across languages, it yields dialogues that lack \textit{naturalness} \citep{lembersky2012language,volansky2015features}, are not properly \textit{localised} nor \textit{culture-specific} \citep{clark2020tydi}. Further, they provide over-optimistic estimates of performance due to the artificial similarity between source and target texts \citep{artetxe2020translation}. As an alternative to translation, new \tod datasets can be created from scratch in a target language through the Wizard-of-Oz framework \citep[WOZ;][]{kelley1984iterative} where humans impersonate both the client and the assistant. However, this process is highly \textit{time- and money-consuming}, thus \textit{failing to scale} to large quantities of examples and languages, and often \textit{lacks coverage} in terms of possible dialogue flows \cite{zhu2020crosswoz,quan2020risawoz}.

% Our contribution: localisation and naturalness/authenticity.
To address all these gaps, in this work we devise a novel \textit{outline-based} annotation pipeline for multilingual \tod datasets that combines the best of both processes. In particular, abstract \textit{dialogue schemata}, specific to individual domains, are sampled from the English Schema-Guided Dialogue dataset \citep[SGD;][]{shah2018building,rastogi2020towards}. Then, the schemata are automatically mapped into outlines in English, which describe the intention that should underlie each dialogue turn and the slots of information it should contain, as shown in Table~\ref{tab:prompt_examples}. Finally, outlines are paraphrased by human subjects into their native tongue and slot values are adapted to the target culture and geography. This ensures both the cost-effectiveness and cross-lingual comparability offered by manual translation, and the naturalness and culture-specificity of creating data from scratch. Through this process, we create the \textbf{C}ross-lingual \textbf{O}utline-based \textbf{D}ialogue dataset (termed \datasetacronym), supporting natural language understanding (intent detection and slot labelling tasks),  dialogue state tracking, and end-to-end dialogue modelling in 11 domains and 4 typologically and areally diverse languages: Arabic, Indonesian, Russian, and Kiswahili.

%% and genealogically 

% VALIDATION
%% assistant's goal orientedness and language fluency, 
To confirm the advantages of the leveraged annotation process, we run a proof-of-concept experiment where we create two analogous datasets through the outline-based pipeline and manual translation, respectively. Based on a quality survey from human participants, we find that, while having similar annotation speed, outline-based annotation achieves significantly higher naturalness and familiarity of concepts and entities, without compromising data quality and language fluency.\footnote{Furthermore, when asked to compare equivalent dialogues obtained with the two processes, respondents favoured outline-based dialogues 8 times out of 10.} Finally, crucial evidence showed that cross-lingual transfer test scores on translation-based data are over-estimated. We demonstrate that this is due to the fact that the distribution of the sentences (and their  hidden representations) is considerably more divergent between training and evaluation dialogues in \datasetacronym than in the translation-based dataset.

%% , whereas those computed on outline-based naturalistic data reveal a much more sobering picture.
%% such as multilingual BERT (mBERT) \citep{devlin2019bert} and XLM on RoBERTa (XLM-R) \citep{Conneau:2020acl} for classification tasks and multilingual T5 (mT5) \citep{mt5} for generation tasks.

% EXPERIMENTS
Further, to establish realistic estimates of performance on multilingual \tod, we benchmark a series of state-of-the-art multilingual \tod models in different \tod tasks on \datasetacronym. Among other findings, we report that zero-shot transfer surpasses `translate-test' on slot labelling, but this trend is reversed for intent detection. Language-specific performance also varies substantially among evaluated models, depending on the quantity of unlabelled data available for pretraining. 

%% (IV) Too specific for intro
%Finally, experimenting with different domains during training and evaluation, we observe an additional gap in performance when models transfer knowledge across domains rather than remaining in-domain.

% CONCLUDING STATEMENT
In sum, \datasetacronym provides the typologically diverse dataset for end-to-end dialogue modelling, and streamlines a scalable annotation process that results in natural and localised dialogues. As such, we hope that \datasetacronym{} will contribute to democratising language technology and facilitating reliable and cost-effective \tod systems for a wide array of languages.  Our data and code are available at \url{https://github.com/cambridgeltl/COD}.

\section{Annotation Design}
\label{sec:design}
%\sparagraph{Motivation} 
\label{ss:motivation}
The main goal of our \tod dataset creation approach is to balance the practical advantages offered by direct translation and the linguistic and cultural specificity granted by bottom-up data collection in the target language. On the one hand, translation of an existing dataset removes the need for a costly and lengthy interactive dialogue generation protocol. By using pre-existing annotated data, dialogue intent labels can be directly transferred to a new language and annotation work is limited to slot value spans. As a consequence, the data are automatically aligned across different languages, which enables direct comparisons of system performance. On the other hand, direct translation is known to perpetuate linguistic and cultural biases into the target language, skewing the syntactic and lexical properties of the data towards the source language, as well as imposing dialogue behaviours and concepts which are not necessarily familiar or appropriate in the target culture. As a result, translated datasets cannot be reliably used as benchmarks of model performance in the target language \citep{koppel2011translationese,volansky2015features,artetxe2020translation,ponti2020xcopa}. 

Our proposed \textit{outline-based} approach aims to marry the benefits of both methods, while avoiding their shortcomings. It achieves time- and cost-effectiveness by bootstrapping from existing dialogue schemata, but refrains from direct translation in favour of outline-guided dialogue writing with target culture-specific slot value adaptation, thus ensuring naturalness and familiarity of the concepts. %and named entities.

%to a linguistic community in question. 

%% (IV): If this is not discussed in Section 3, return these sentences there
%%The results of the comparative qualitative analysis reported in Section~\ref{s:qualitative} demonstrate that our outline-based approach indeed satisfies these objectives, improving upon direct translation with respect to naturalness and concept familiarity, while maintaining the latter method's benefits in terms of cross-lingual data alignment and annotation transfer. 

\rparagraph{Source Data}
We selected the English Schema-Guided Dialogue (SGD) dataset \citep{shah2018building,rastogi2020towards} as our starting point due to its scale (20k human-assistant dialogues) and diversity (20 different domains). The SGD dataset construction paradigm combined automatic generation of dialogue schemata and manual creation of dialogue paraphrases by crowdworkers. The method, dubbed ``Machines Talking To Machines'' (M2M), is an alternative to the popular human-to-human Wizard-of-Oz framework \citep{kelley1984iterative}, where pairs of crowdworkers interact following task specifications, generated through sampling of slot values from an API client, in order to complete a certain goal and their conversations are directly recorded \citep{wen2016multi,budzianowski2018multiwoz}. The crowdsourced dialogues then undergo another round of annotation with dialogue acts and slot spans. While the WOZ approach has the advantage of collecting actual human-to-human conversations, the process is expensive and prone to error, given the risk that the free-form interactions might not exhaustively cover possible interactions or might not lend themselves to direct use for model training (e.g., long and too convoluted exchanges). %that are too long and convoluted).

The SGD's M2M approach has the advantage of greater speed and cost-effectiveness. In the first stage, it simulates the user-assistant interaction to exhaustively explore possible user behaviours and dialogue scenarios and generate dialogue outlines (i.e., template utterances and their semantic parses), maximising diversity and coverage of different dialogue flows by means of permutations of slots, intents and domains. Subsequently, crowdworkers are tasked with paraphrasing dialogue templates to create natural language (NL) utterances, preserving the meaning and key elements captured in the templates (e.g., outline: ``Book movie with title is Inside Out and date is tomorrow'' $\rightarrow$ paraphrase: \textit{I want to buy tickets for Inside Out for tomorrow}.), and subsequently validate slot spans. Given that dialogue intents and slot values are provided in the dialogue outlines, the risk of erroneous labels in the final dataset is minimised.

The SGD dataset organises dialogue data as lists of turns for each individual interaction, each turn containing an utterance by the user or system. The accompanying annotations are grouped into frames, each corresponding to a single API or service (e.g., \textit{Banks\_2}). In turn, each service is represented as a schema, i.e., a normalised representation of a service-specific interface, and includes its characteristic functions (intents) and parameters (slots), as well as their NL descriptions.\footnote{For example, the ``\textit{Alarm\_1}'' service comprises intents such as ``\textit{GetAlarms}'' (``\textit{Get the alarms user has already set}'') and ``\textit{AddAlarm}'' (``\textit{Set a new alarm}'') and slots ``\textit{alarm\_time}'', ``\textit{alarm\_name}'', ``\textit{new\_alarm\_time}'' and ``\textit{new\_alarm\_name}''.}
    
%% Moved to Section 3
%For the purposes of a comparative evaluation of two dialogue creation methods, (i) direct translation and (ii) prompt-based generation, we used the English SGD data in two ways. In (i), randomly sampled (see Section \ref{ss:protocol}) English user/system utterances were extracted directly from the dataset with accompanying slot and intent annotations and subsequently translated into the target language by professional translators, also responsible for validating target language slot spans. In (ii), we automatically extracted dialogue frames, including intents and slots, corresponding to the dialogue IDs sampled in (i), and used them to generate natural language prompts to guide manual dialogue creation by target language native speakers. We explain this process in detail in Section \ref{ss:protocol}.

\iffalse
pros: 
(i) diversity of language and dialogue flows, 
(ii) coverage of all expected user behaviours, 
(iii) correctness of supervision labels. 
(iv) boot-strapping dialogue agents so that they can be deployed to serve real users with an acceptable task completion rate, after which they should be improved directly from user feedback using reinforcement learning.
\fi

\begin{table*}[!t]
\def\arraystretch{0.83}
{\footnotesize
\begin{tabularx}{1.0\textwidth}{llXXXrr}
\toprule
\textbf{Language} & \textbf{\textsc{iso}} & \textbf{Family} & \textbf{Branch} & \textbf{Macro-area} & \textbf{L1} [M] & \textbf{Total} [M]\\
\cmidrule(lr){3-7}
Russian & \textsc{ru} & Indo-European & Balto-Slavic & Eurasia & 153.7 & 258 \\
Standard Arabic & \textsc{ar} & Afro-Asiatic & Semitic & Eurasia / Africa & 0$^\dagger$ & 274\\
Indonesian & \textsc{id} & 	Austronesian & Malayo-Polynesian & Papunesia & 43.6 & 199\\
Kiswahili & \textsc{sw} & Niger–Congo & Bantu & Africa & 16.3 & 69\\
 
\bottomrule
\end{tabularx}
}%
\vspace{-1mm}
\caption{Language stats. The last two columns denote the number of speakers. $^\dagger$Standard Arabic is learned as L2.}
\label{tab:languages}
\vspace{-1.5mm}
\end{table*}

\begin{table*}[!t]
\centering
%% Further information about the datasets is provided in Appendix~\ref{app: multilingual datasets comparisons}
%\resizebox{}{1.0\textwidth}{
\def\arraystretch{0.8}
{\footnotesize
\begin{tabular}{lcccccccc} 
\toprule
\rowcolor{Gray}
& \multicolumn{5}{c}{\bf NLU-Only Datasets} & \multicolumn{2}{c}{\bf End-to-End Datasets} & \\
\cmidrule(lr){2-6} \cmidrule(lr){7-8}
             & \multicolumn{1}{l}{M. TOP} & \multicolumn{1}{l}{M. ATIS} & \multicolumn{1}{l}{MultiATIS++} & \multicolumn{1}{l}{MTOP} & \multicolumn{1}{l}{xSID} & BiTOD & GlobalWOZ & \textbf{\cod}  \\ 
%{} & {\scriptsize \citet{schuster2019cross}} & {} & {\scriptsize \citet{xu2020multiatis++}} & {\scriptsize \citet{li2020mtop}} & {} & {\scriptsize \citet{lin2021bitod}} & {\scriptsize \citet{ding2021globalwoz}} & {\scriptsize Ours} \\
\cmidrule(lr){2-9}
\# languages & 3                          & 3                           & 9                               & 6                        & 13                       & 2     & 3         & 4              \\
Typology     & 0.20                       & 0.29                        & 0.33                            & 0.29                     & 0.37                     & 0.15  & 0.24      & 0.31           \\
Family       & 0.67                       & 0.67                        & 0.44                            & 0.33                     & 0.50                     & 1.0   & 0.75      & 1.0            \\
Macroareas   & 0                          & 0                           & 0                               & 0                        & 0.26                     & 0     & 0.14     & 1.04           \\
\bottomrule
\end{tabular}
}%
\vspace{-0.5mm}
\caption{Comparison of diversity indices of multilingual dialogue datasets in terms of typology, family, and macroareas. For the description of the three diversity measures, we refer the reader to \citet{ponti2020xcopa}. M. TOP was created by \citet{schuster2019cross}; M. ATIS \citep{upadhyay2018multiatis}; MultiATIS++ \citep{xu2020multiatis++}; MTOP \cite{li2020mtop}; xSID \citet{van-der-goot-etal-2021-masked}; BiTOD \cite{lin2021bitod}; GlobalWOZ \cite{ding2021globalwoz}.}
\label{tab:language diversity}
\vspace{-2mm}
\end{table*}

\begin{table}[!t]
\def\arraystretch{0.73}
{\footnotesize
\begin{tabularx}{1.0\textwidth}{lXX}
\toprule
\textbf{Domain} & \textbf{Dev} & \textbf{Test}\\
 \cmidrule(lr){2-3}
 Alarm ($\diamondsuit$) & 13 & 21\\
 Flights & 12 & 23\\
 Homes & 12 & 13 \\
 Movies & 16 & 19 \\
 Music & 14 & 16\\
 Media & - & 17\\
 Banks & 14 & -\\
 Payment ($\diamondsuit$) & - & 8\\
 RideSharing & - & 11 \\
 Travel & 12 & -\\
 Weather & 18 & -\\
 \cmidrule(lr){2-3}
\textbf{\# dialogues} & 92  & 102 \\
\textbf{\# turns} & 1138 & 1352 \\
\bottomrule
\end{tabularx}
}%
\vspace{-0.5mm}
\caption{Number of dialogues per domain in the development and test set and dataset statistics. $\diamondsuit$ marks the domains which are not included in the training set.}
\label{tab:domains}
\vspace{-2mm}
\end{table}

%% dialogue generation
\rparagraph{Languages}
To assess the viability of the outline-based method, we selected Russian as a trial language and carried out data collection using two methods: (i) direct translation from English and (ii) our proposed outline-based approach. Having evaluated the quality of the output of both methods and the advantages of in-target outline-based creation (see later \S\ref{s:qualitative}), we applied the method to three other languages which boast a large number of speakers and yet suffer from a shortage of resources: Arabic, Indonesian, and Kiswahili, ensuring the dataset's diversity in terms of language family (Indo-European (\textsc{ru}), Afro-Asiatic (\textsc{ar}), Austronesian (\textsc{id}) and Niger-Congo (\textsc{sw})) and macro-area (Eurasia, Papunesia, Africa), as well as writing systems (Cyrillic, Arabic, and Latin scripts). 

We present a quantitative evaluation of the linguistic diversity of the language sample in Table~\ref{tab:language diversity}. We also compare it with the standard multilingual dialogue NLU and end-to-end datasets. In terms of typology, \cod is comparable to others which have much larger language samples (e.g., MultiATIS++ or xSID) and considerably exceeds others. With respect to family and macroarea diversity, \cod is the most diverse out of existing datasets.

\begin{table*}[!t]
\def\arraystretch{0.87}
{\footnotesize
\begin{tabularx}{0.99\textwidth}{lXXlX}
\toprule
\textbf{Act} & \textbf{Slot/Intent} & \textbf{Description} & \textbf{Value} & \textbf{Outline}\\
\cmidrule(lr){2-5}
INFORM\_INTENT & SearchOnewayFlight & Search for one-way flights to the destination of choice & -- & \textit{\textbf{Express the desire to} search for one-way flights} \\
\cmidrule(lr){2-5}
 REQUEST & number\_checked\_bags & Number of bags to check in & 2 & \textit{\textbf{Ask if} the number of bags to check in is 2} \\
 
\bottomrule
\end{tabularx}
}%
\vspace{-0.5mm}
\caption{Examples of dialogue generation outlines created from SGD schemata, that is, annotations of dialogue acts, intents, slots and values, with intent-specific rewrites in bold.}
\label{tab:prompt-rewrites}
\vspace{-1.5mm}
\end{table*}

\subsection{Annotation Protocol}
\label{ss:protocol}
The data creation protocol involved the following phases: \textbf{1)} source dialogue sampling, \textbf{2)} automatic generation of outlines based on intent and slot information using rewrite rules, \textbf{3)} manual outline-driven target language dialogue creation and slot annotation, \textbf{4)} post-hoc review, described here. %in what follows

\rparagraph{Source Dialogue Sampling}
To ensure wide coverage of dialogue scenarios, we randomly sampled source dialogues from across 11 domains, out of which five (\textit{Alarm, Flights, Homes, Movies, Music}) are shared between the development and test set; the remainder are unique to either set, to enable \textit{cross-domain} experiments. To guarantee a balanced coverage of different intents, we sampled 10 examples per intent, which ensures the task cannot be solved by simply predicting the most common intent. Table~\ref{tab:domains} summarises the coverage of domains and the number of dialogues and turns as a result of this sampling procedure.

\rparagraph{Outline Generation} The goal of this step is to create minimal but sufficient instructions for target language dialogue creators to ensure coverage of specific intents and slots, while avoiding imposing predefined syntactic structures or linguistic expressions. First, for each user or system act, we manually create a rewrite rule, e.g.,  REQUEST\_ALTS$\rightarrow$\textit{Request alternative options} or INFORM\_COUNT$\rightarrow$\textit{Inform the user that you found} + \texttt{INFORM\_COUNT[value]} + \textit{such option(s)} (where \texttt{value} corresponds to the number of options matching the user request). Next, we automatically match each intent and slot with its NL description (provided in the SGD schemata) and used them to generate intent/slot-specific outlines (with stylistic adaptations where necessary): e.g., an intent ``SearchOnewayFlight'' and a description ``\textit{Search for one-way flights to the destination of choice}'' would yield an outline \textit{Express the desire to search for one-way flights} (see Table~\ref{tab:prompt-rewrites}). 

\rparagraph{Dialogue Writing}
We recruited target language native speakers fluent in English via the \url{proz.com} platform.\footnote{To ensure quality, we restricted the candidate pool to users with reported target language credentials and relevant experience. Successful candidates were selected based on a qualification exercise consisting in writing a 6-turn dialogue according to outlines, analogous to those in the main task.} Dialogue creators were presented with language-specific dialogue creation guidelines (see Appendix~\ref{app:guidelines}), which described the goals of the task, i.e., creative writing of natural-sounding exchanges between a hypothetical user and a \tod system. An essential part of the task consisted in a cultural adaptation of the slot values, illustrated in Table~\ref{tab:prompt_examples}. For all culturally and geographically specific slot values (e.g., city names, movie titles, names of artists), creators were asked to substitute them with named entities more familiar or closer to their culture (e.g., American Airlines$\rightarrow$Aeroflot, New York$\rightarrow$Jakarta). 

%% (IV, removed, this is redundant)
%%In \S\ref{s:qualitative}, we summarise the results of a qualitative analysis of the data output by this method compared to direct translation without slot value substitutions in terms of perceived familiarity by target language native speakers.  

\rparagraph{Slot Span Validation}
Creators performed the first round of slot span labeling while working on dialogue writing. Subsequently, the annotated data in each language underwent an additional round of manual revision by a target language native speaker and a final automatic check for slot value-span matches. We verified inter-annotator reliability on slot span labeling on Russian, where we collected slot span annotations from pairs of independent native-speaker annotators. The obtained accuracy scores (i.e., ratio of slot instances with matching spans to the total annotated instances) of 0.99 for dev data and 0.98 for test data reveal very high agreement on this task.

%\section{Qualitative and Quantitative Analyses}
\section{Translation versus Outline-Based}
\label{s:qualitative}
%\subsection{Translation vs Outline-based Generation}

%% to dialogue dataset creation was 
The main motivation behind the outline-based approach is to avoid the known pitfalls of \textit{direct translation} and produce evaluation data better representing the linguistic and cultural realities of each language in the sample (see \S\ref{ss:motivation}). To verify whether the method satisfies these goals in practice, we first carried out a trial experiment consisting in parallel dialogue data creation using two different methods, (i) direct translation and (ii) outline-based generation. To ensure a fair comparison, we used the same sample of source SGD dialogues in both tasks, in two different ways. In (i), randomly sampled (see \S\ref{ss:protocol}) English user/system utterances were extracted directly from the dataset with accompanying slot and intent annotations and subsequently translated into the target language by professional translators, also responsible for validating target language slot spans. In (ii), we automatically extracted dialogue frames, including intents and slots, corresponding to the dialogue IDs sampled in (i), and used them to generate NL outlines to guide manual dialogue creation by native speakers, relying on the procedure described in \S\ref{ss:protocol}.

%To compare the time needed to carry out dialogue generation or translation, with slot annotation, 

We also asked the participants to time themselves while working on the task. Notably, we found the annotation speed to be identical for the two methods, averaging 15 seconds per single dialogue turn (dialogue writing + slot annotation). While the translation approach does not require any creative input in terms of cultural adaptations of slot values, the outline-based approach allows freedom in terms of the linguistic expressions used, removing the need for faithful translation of the original English sentences, which ultimately results in similar time requirements on both tasks.

\rparagraph{Quality Survey}
To compare the two methods' output, we carried out a quality survey with 15 Russian native speakers. It consisted of two consecutive parts: (1) independent and (2) comparative evaluation; the non-comparative part came first so as to avoid priming effects from an a priori awareness of systematic qualitative differences between examples coming from either method. Within each part, the order of questions was randomised. In Part 1, the respondents were presented with 6 randomly sampled dialogues from the data generated by either method (3 dialogues per method) and were asked to answer to what extent they agree with each of four statements (provided in Table~\ref{tab:survey}) by giving a rating on a 5-point Likert scale. In Part 2, respondents were presented with 5 randomly sampled pairs of matching dialogue excerpts (i.e., a set of $N$ dialogue turns extracted based on a shared dialogue ID from both datasets) and were asked to choose which excerpt (A or B) sounded more natural to them. All survey questions and instructions were translated into Russian.
%% rather than via translation from English, 

\begin{figure}[!t]
  %\begin{center}
    
    \includegraphics[width=0.93\textwidth]{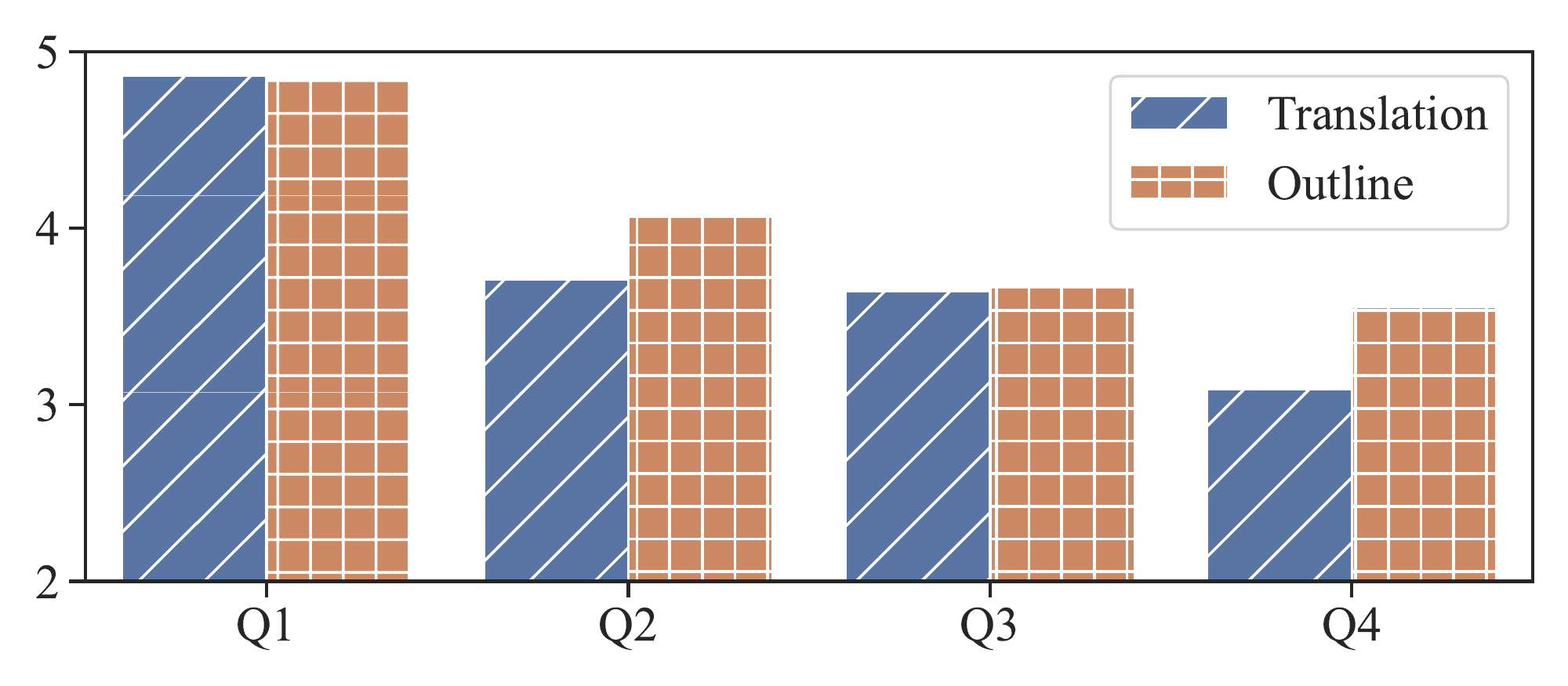}
  %\end{center}
  \vspace{-1mm}
  \caption{Average scores for each quality survey question (see Table~\ref{tab:survey}) assigned to dialogue examples generated via \textit{translation} versus \textit{outline}-based generation.}
  \label{fig:part1}
  \vspace{-2mm}
\end{figure}

Figure~\ref{fig:part1} shows average scores for each question in Part 1 across all 15 participants. The methods produce dialogues which score very similarly in terms of the assistant's goal-orientedness (Q1) and Russian language fluency (Q3). However, the differences are clearer in scores for Q2 and Q4. First, the user utterances created based on outlines are perceived as more natural-sounding (Q2). Further, they score noticeably better in terms of the familiarity of mentioned entities. These results are encouraging, given that Q4 directly addresses one of the main objectives of our method, i.e., target language-specificity.  While both approaches are capable of producing convincing Russian dialogues, the results of Part 2 are more clearly skewed in favour of the outline-based method: out of 75 comparisons (15 participants judging 5 pairs each), outline-based dialogues are preferred (i.e., judged as more natural-sounding) in 80\% of cases. In Table~\ref{tab:comparison} we show an example pair of dialogue excerpts from each method, analogous to those used in the survey, with accompanying English translations.

\begin{table}[!t]
\def\arraystretch{0.9}
{\footnotesize
\begin{tabularx}{1.0\textwidth}{X}
\toprule
\rowcolor{Gray}
\textbf{Instructions} \\
\midrule
\textit{Please state to what extent you agree/disagree with each statement on the scale of 1-5 (1-Strongly disagree, 5-Strongly agree)} \\
\midrule
\rowcolor{Gray}
\textbf{Questions}\\
\midrule
Q1. The ASSISTANT helps satisfy the USER’s requests.\\
Q2. The USER speaks naturally and sounds like a Russian native speaker.\\
Q3. The ASSISTANT speaks naturally and sounds like a Russian native speaker.\\
Q4. I can easily imagine myself mentioning or hearing the proper names referred to in the dialogue (e.g., titles of films or songs, people, places) in a conversation with my Russian friends or family.\\

\bottomrule
\end{tabularx}
}%
\vspace{-0.5mm}
\caption{Quality survey questions (Part 1).}
\label{tab:survey}
\vspace{-1.5mm}
\end{table}

\begin{table*}
{\footnotesize
\def\arraystretch{0.83}
\begin{tabularx}{1.0\textwidth}{XX}
\toprule
\textbf{Translation} &  \textbf{Outline-based Generation}\\
\cmidrule(lr){1-1} \cmidrule(lr){2-2}

USER: \foreignlanguage{russian}{Можете ли вы найти мне квартиру, в которой можно держать домашних животных?} \hspace*{0pt}\hfill $\clubsuit$ &	USER: \foreignlanguage{russian}{Мне нужно найти квартиру в аренду в которой можно селиться с животными.} \\
\rowcolor{grey}
\textsc{en}: \textit{Can you find me an apartment that allows pets?} & \textit{I need to find a flat to rent in which I can live with pets.} \\
ASSISTANT: \foreignlanguage{russian}{Alma Apartments предлагают 1 спальню с 1 ванной комнатой за 1 900\$ в месяц. <...>}	& ASSISTANT: \foreignlanguage{russian}{Позволь мне предложить следующий вариант. Это однокомнатная квартира с одной ванной комнатой. Квартира находится в жилом комплексе Белая вишня. <...>} \\
\rowcolor{grey}
\textsc{en}: \textit{Alma Apartments has a 1 bedroom 1 bathroom for 1,900 per month. <...>} & \textit{Let me propose the following option. It is a 1 bedroom 1 bathroom apartment. The apartment is located in the "White Cherry" building. <...>}\\
USER: \foreignlanguage{russian}{Звучит многообещающе. Я бы хотел взглянуть на неё. Не могли бы вы назначить визит на 10-е число? <...> Если подумать, то лучше 12-го числа этого месяца.}	& USER: \foreignlanguage{russian}{Отлично. Можно договориться о просмотре квартиры 10-го числа? <...> Ой не, лучше на 12 число.} \\
\rowcolor{grey}
\textsc{en}: \textit{Sounds promising. I'd like to look it over. Could you schedule a visit on the 10th? <...> On second thought, make it the 12th of this month.} & \textit{Perfect. Could we arrange the visit on the 10th? <...> Oh no, better on the 12th.}\\
ASSISTANT: \foreignlanguage{russian}{Итак, вы хотите посетить Alma Apartments 12 марта.
} &	ASSISTANT: \foreignlanguage{russian}{Резервирую осмотр квартиры в жилом комплексе Белая вишня на 12 марта?} \\
\rowcolor{grey}
\textsc{en}: \textit{Okay, you would like to visit Alma Apartments on March 12th.} &
\textit{Am I booking the visit to "White Cherry" on the 12th of March?}\\
USER: \foreignlanguage{russian}{Да. Какой там номер телефона?}	& USER: \foreignlanguage{russian}{Да, именно так. Можешь найти контактный номер телефона?} \\
\rowcolor{grey}
\textsc{en}: \textit{Yes. What is their phone number?} & \textit{Yes, exactly. Could you find their contact number?}\\
ASSISTANT: \foreignlanguage{russian}{650-813-1369. Ваш визит запланирован.
} \hspace*{0pt}\hfill $\clubsuit \spadesuit$ &	ASSISTANT: \foreignlanguage{russian}{Да, номер телефона 650-813-1369. Я успешно забронировала осмотр.} \\
\rowcolor{grey}
\textsc{en}: \textit{It's 650-813-1369. Your visit is scheduled.} & \textit{Yes, the phone number is 650-813-1369. I successfully booked the visit.} \\

\bottomrule
\end{tabularx}}
\vspace{-0.5mm}
\caption{Comparison of dialogues generated by each method. For each user/assistant utterance, we provide the original English sentences from SGD for the translation method, and English translations of the Russian utterances written based on outlines.  $\clubsuit$ -- syntactic similarity to source language; $\spadesuit$ -- lexical similarity to source language.}
\label{tab:comparison}
\vspace{-1.5mm}
\end{table*}

\rparagraph{Effects of \textit{Translationese}} 
\label{ss:translationese}
% One of the motivations for the outline-based generation method was to avoid the effects of "translationese" \citep{koppel2011translationese}, i.e., direct translation is often skewed towards lexical and syntactic choices which are more typical of the source language than the target. We present several examples of such influences in . 
Dialogue data are expected to be representative of a natural interaction between two interlocutors. %This means that 
The utterances of both the user \textit{and} the system should reflect the properties characteristic of the conversational register in a given language, appropriate for the communicative situation at hand and the participants' social roles \citep{chaves2019register,chaves2021should}.   %and be informal and colloquial.
%%OM: I'd avoid 'informal', considering that in some cultures/languages an exchange with a stranger may require a level of formality/politeness
When qualitatively comparing the translation and outline-based generation in Table~\ref{tab:comparison}, we observe that translated utterances are often skewed to the source language syntax and lexicon (known as the ``translationese'' effects, \citealt{koppel2011translationese}), compromising fluency and idiomacity that are essential in natural-sounding exchanges.  %This can withhold it from sounding natural and colloquial in the target language.

%Starting from the syntactic patterns, 
One issue which arises in literal translation is syntactic calques from the source language. For instance, the translation of the first USER utterance (Table~\ref{tab:comparison}, col. `Translation') uses a dative pronoun \foreignlanguage{russian}{найти мне}\textsc{[dative]} (\textit{find me}), even though the transitive verb \foreignlanguage{russian}{найти} (\textit{find}) does not require the \textsc{[dative]} case after it -- a likely calque of the English expression \textit{Can you find me}. In comparison, the corresponding outline-based generated utterance uses a more fluent construction.  Another problem concerns the differences in the use of grammatical structures depending on the language register. For instance, using passive voice in spoken English is common (cf. last ASSISTANT utterance in Table~\ref{tab:comparison}). The literal translation of the dialogue into Russian also includes passive voice, although it is usually avoided in spoken Russian \citep{babby1975syntax}. In contrast, the outline-based utterance uses a simpler active voice construction which has the same meaning as the one in the translation.  

We observe further ``translationese'' effects on the lexical level, namely (i) the preference for lexical cognates of source language words, and (ii) the use of a vocabulary typical for the written language; both are exemplified by the last ASSISTANT utterance (Table~\ref{tab:comparison}). The translation includes the verb \foreignlanguage{russian}{запланирован} (\textit{is planned}), even though the verb \foreignlanguage{russian}{планировать}, having the same root as English \textit{to plan}, is rarely used in spoken Russian with regards to arranging near-future appointments and more frequently with regards to making a step-by-step plan. In contrast, the outline-based generated utterance includes the verb  \foreignlanguage{russian}{забронировать} (\textit{to book}) which is more specific to arranging appointments and more frequently used in spoken language. Similar examples for both (i) and (ii) are presented in Appendix~\ref{app: examples calques}.

\rparagraph{Evaluation of \tod Systems on Translation-Based versus Outline-Generated Data} 
The vast majority of existing NLU datasets is based on translation from English to the target language \citep{xu2020multiatis++, van-der-goot-etal-2021-masked}. This could lead to overly optimistic evaluation of cross-lingual \tod systems as the translations might not be representative of users' language use in real life. We expect that translation-based evaluation data will lead to overly optimistic performance as it suffers from the effects of ``translationese'' demonstrated above.

In this diagnostic experiment, we use a \textit{translate-train} approach where: (i) training data are translated from the source language (\texttt{en}) to the target (\texttt{ru}) via Google Translate; and (ii) the model is fine-tuned on these automatically translated data. In our analysis we test the model on evaluation data obtained in each of the following ways: \textbf{(a)} translated using Google Translate, \textbf{(b)} translated by professional translators (closest in nature to existing dialogue NLU datasets), \textbf{(c)} generated based on outlines.\footnote{We focus on the intent detection task to avoid the interference of noise introduced by the alignment algorithms (i.e., aligning the source language examples with automatic translations of the training data for slot labelling).} For the experiment, we fine-tune mBERT \cite{devlin2019bert} on intent detection.\footnote{A summary of training hyperparameters is provided later in \S\ref{s:experiments} and in Appendix~\ref{app:training hyperparameters}.}

%We hypothesise that test performance on a) and b) will be high and will drop on c), as both a) and b) are translation-based while c) is expected to present the model with more complex, more natural examples.  

The results in Table~\ref{table: diff test datas analysis} indicate that the stronger performance is observed on translation-based evaluation sets than on more natural, outline-based generated examples. The results corroborate previous observations in other areas of NLP, e.g., machine translation \cite{graham2020statistical}, now for \tod. Crucially, this experiment verifies that using solely translation-based \tod evaluation data might yield an overly optimistic estimation of models' cross-lingual capabilities and, consequently, too optimistic performance expectations in real-life applications. This further validates our proposed outline-based approach to (more natural and target-grounded) multilingual \tod data creation.

\begin{table}[!t]
\centering
\def\arraystretch{0.83}
{\footnotesize
\begin{tabularx}{\columnwidth}{l XX} 
\toprule
\textbf{Data Creation}                                                               & \textbf{Split} & \textbf{Accuracy}  \\ 
\cmidrule(lr){2-3}
\multirow{2}{*}{\begin{tabular}[c]{@{}l@{}}Google\\Translate\end{tabular}}                                                    & Dev        & 47.98        \\
                                                                                     & Test       & 35.06        \\ \cmidrule(lr){2-3}
\multirow{2}{*}{\begin{tabular}[c]{@{}l@{}}Professional\\Translation\end{tabular}} & Dev        & 48.33        \\
                                                                                     & Test       & 34.62        \\ \cmidrule(lr){2-3}
\multirow{2}{*}{\begin{tabular}[c]{@{}l@{}}Outline-based\\Generation\end{tabular}}    & Dev        & 40.25        \\
                                                                                     & Test       & 31.81        \\
\bottomrule
\end{tabularx}
}%
\vspace{-0.5mm}
\caption{Cross-lingual intent detection accuracy on dev and test data coming from three different data creation methods: (\textbf{a)} translated via Google Translate; (\textbf{b)} translated by professionals; and \textbf{(c)} outline-based generation.}
\label{table: diff test datas analysis}
\vspace{-1mm}
\end{table}

\rparagraph{Analysis of Sentence Encodings}
\label{ss:encodings}
%% Genie? :)
% The vast majority of existing NLU datasets is based on translation from English to the target language \citep{xu2020multiatis++, van-der-goot-etal-2021-masked}. This could lead to overly optimistic evaluation of NLU systems as the translations might not be representative of users' language use in real life. 
%As shown above, evaluating the model only on translation based dialogue data can give an overly optimistic impression of the model's cross-lingual capabilities. 
One reason behind the scores observed in Table~\ref{table: diff test datas analysis} might lie in the differences between multilingual sentence encodings of English examples, examples generated via translation, and examples generated via the outline-based approach. To test this, we obtain sentence encodings of all user turns for one intent from the three datasets via the distilled multilingual USE sentence encoder \citep{yang:muse,reimers-2019-sentence-bert}.\footnote{The same trends in the results were observed with other standard multilingual sentence encoders such as LaBSE \cite{feng2020language}, see Appendix~\ref{app:cosine similarities encodings} for additional results.}

%\footnote{Available at \url{https://huggingface.co/sentence-transformers/distiluse-base-multilingual-cased-v1}}. The model supports 15 languages, including Russian and English. 

\begin{figure}[!t]
  %\begin{center}
    \centering
    \includegraphics[width=0.75\textwidth]{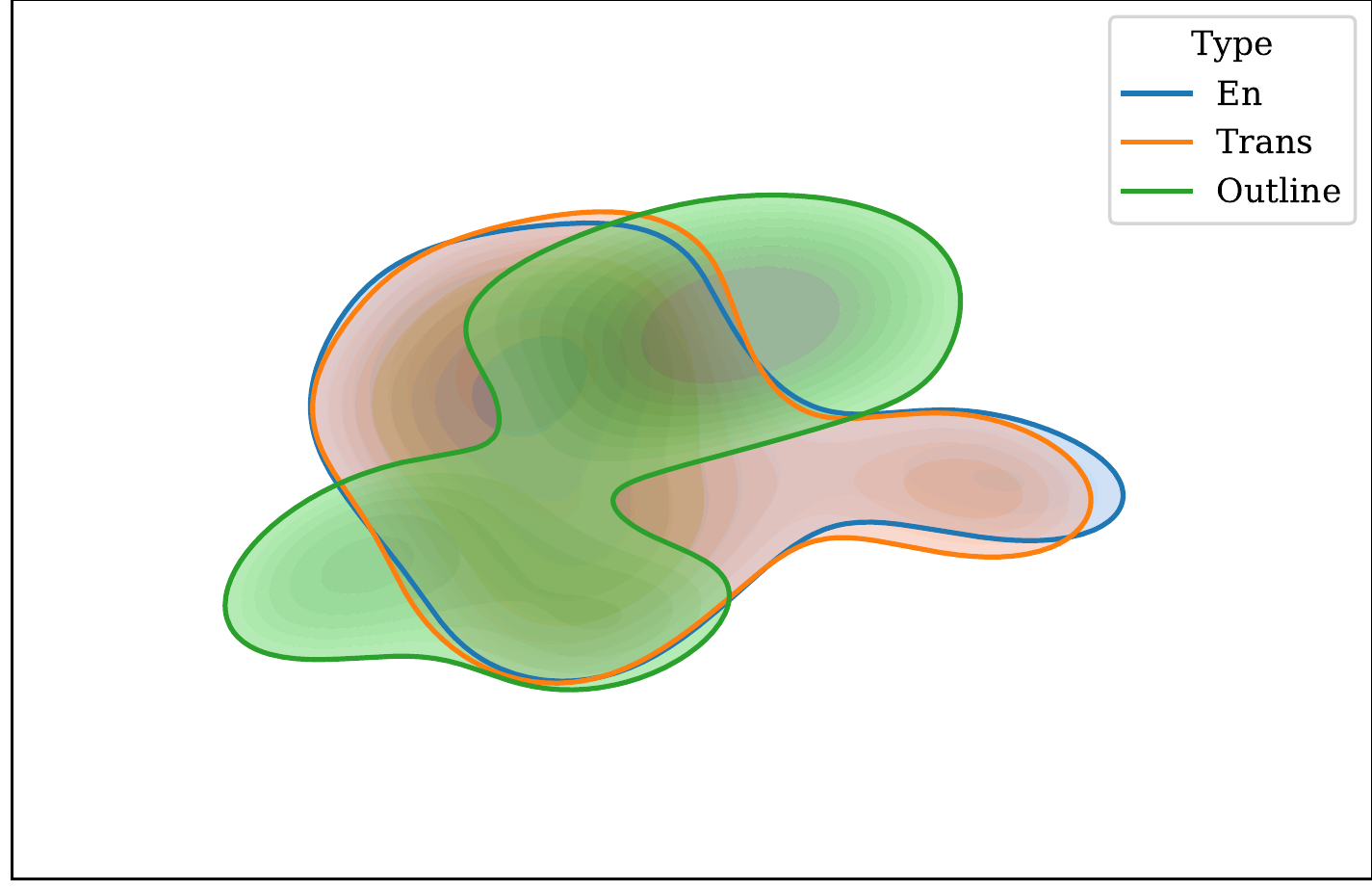}
  %\end{center}
  \caption{Kernel density estimate (KDE) plot for distributions of representations obtained by encoding user turns via distilled multilingual USE as the sentence encoder. The input sentences are either the original sentences in English (\texttt{En}), translated to Russian (\texttt{Trans}), or generated in Russian based on Outlines (\texttt{Outline}). Dimensionality reduction was performed using tSNE \citep{van2012visualizingtsne}. Pairwise KL-divergence scores between KDE-estimated Gaussians: KL (En || Trans) = $7.5 \times 10^{-4}$; KL (En || Outline) = $4.69 \times 10^{-5}$; KL (Trans || Outline) =  $3.84 \times 10^{-5}$.}
  \label{fig: tsne translations}
\end{figure}

As illustrated in Figure~\ref{fig: tsne translations}, the translation-based data are encoded into sentence representations that are much more similar to their English source than the corresponding outline-generated examples. The difference holds across dev and test splits and across different multilingual sentence encoders (see also Appendix~\ref{app:cosine similarities encodings}). This indicates that, as expected, the utterances obtained via translation are artificially more similar to their English counterparts than the outline-generated ones. This again underlines the finding from Table~\ref{table: diff test datas analysis}: multilingual \tod datasets collected via outline-based generation should lead to more realistic assessments of multilingual \tod models than translation-based multilingual \tod datasets.

%% This the cross-lingual transfer performance of any model based on state-of-the-art pretrained multilingual models will be overly optimistic when assessed on translation-based datasets. 

% Genie: prev KDE plot and this one look differently due to different random seeds.

\rparagraph{Further Discussion}
% pros and cons; ultimate goal: language-specific dialogue scenarios representative of a given culture, rather than taken from an English source
To meet the urgent, ever-growing demand for large-scale multilingual \tod datasets, data collection methods which efficiently leverage existing resources to generate new data fast without compromising data quality are especially needed. Direct translation has the benefit of re-using already annotated and verified data entries, moreover, it is a well-defined task which does not require task-specific guidelines or training. However, as we demonstrated here, it unnaturally skews the data towards the source language. This makes evaluation results unreliable. Our proposed `bottom-up' approach produces a much more realistic benchmark for gauging models’ multilingual capabilities. The outlines provide minimal instructions to annotators, which, together with the task guidelines, prove sufficient for a single annotator to create natural-sounding user-assistant exchanges capturing predefined intents and slot values. This circumvents the need for setting up an expensive WOZ pipeline, where pairs of users interact live and their conversations are recorded.

%% improves upon the translation method by eliciting language-specific utterances created in the target language bottom-up, based on outlines, thus 

%% From the perspective of language representativeness, 

An important area for future improvement concerns cultural debiasing of the topics, situations and concepts captured in the dialogues. Although our method generates linguistic expressions which are natural in the target language, the dialogue scenarios included in the dataset are still inherited from English. While most of these are common around the globe (e.g., searching for a property to rent, selecting a movie to watch), some are much less likely to happen in some cultures or communities (e.g., making a public money transfer). Looking ahead, creating dialogue technology and resources that are representative of and applicable within individual  communities of speakers should involve a careful selection of dialogue scenarios, based on their relevance and plausibility in the culture in question, as very recently started in other NLP areas \cite[e.g.,][]{liu-etal-2021-visually}. In our current dataset, we ensured the applicability and comprehensibility of the concepts referred to in the dialogues by entrusting native speakers with cultural adaptations and replacements of foreign concepts with those common in their culture and environment.

\section{Baselines, Results, Discussion}
\label{s:experiments}
\datasetacronym{} includes labelled data and thus enables experimentation for three standard \tod tasks: i) Natural Language Understanding (NLU; intent detection and slot labelling); ii) dialogue state tracking (DST); and iii) end-to-end (E2E) dialogue modelling. In what follows, we benchmark a representative selection of state-of-the-art models (\textsection \ref{subsec: baselines}) on our new dataset, highlighting its potential for evaluation and the key challenges it presents across different tasks and experimental setups (\textsection \ref{subsec: results}).   %and highlight and to highlight which aspects of the dialogue systems can be tested with it. %In \textsection \ref{subsec: baselines} we describe the architectures for the baseline models we use in our experiments. In \textsection \ref{subsec: results}, we present and discuss the results. 

\rparagraph{Notation} A dialogue $\mathcal{D}$ is a sequence of alternating user and system turns $\{\mathcal{U}_1, \mathcal{S}_1, \mathcal{U}_2, \mathcal{S}_2, ...\}$. Dialogue history at turn $t$ is the set of turns up to point $t$, i.e., $\mathcal{H}_t = \{\mathcal{U}_1, \mathcal{S}_1, ..., \mathcal{U}_{t-1}, \mathcal{S}_{t-1}, \mathcal{U}_{t}\}$.

\begin{table*}[!t]
\centering
\resizebox{\columnwidth}{!}{%
{\footnotesize
\def\arraystretch{0.87}
\begin{tabular}{lll;{1pt/1pt}lllll;{1pt/1pt}lllll} 
\toprule
\rowcolor{Gray}
\multicolumn{3}{l;{1pt/1pt}}{}                      & \multicolumn{5}{c;{1pt/1pt}}{\bf Intent Detection} & \multicolumn{5}{c}{\bf Slot Labelling}     \\
\rowcolor{Gray}
\textbf{Setup}                         & \textbf{TrSystem} & \textbf{Model} & \textsc{AR}    & \textsc{ID}    & \textsc{RU}    & \textsc{SW}    & AVG                 & \textsc{AR}    & \textsc{ID}    & \textsc{RU}    & \textsc{SW}    & AVG    \\ 
\cmidrule(r){1-13}
MEncoder                        &          & mBERT & 18.61 & 17.57 & 22.83 & 6.09  & 16.28               & 21.54 & 15.29 & 24.89 & 8.84  & 17.64  \\
\hdashline
\multirow{2}{*}{TrTest} & GTr      & mBERT & 24.46 & 27.34 & 28.97 & 23.93 & 26.18               & 11.70 & 16.36 & 19.56 & 16.67 & 16.07
  \\
                                 & MarianMT   & mBERT & 28.40 & 26.89 & 29.38 & 25.38 & 27.51               &  13.28 & 14.89 & 20.21 & 11.98 & 15.09 \\ 
\cmidrule(r){1-13}
MEncoder                        &          & XLM-R & 25.56 & 29.88 & 27.60 & 19.59 & 25.66               & 28.65 & 31.73 & 32.47 & 15.18 & 27.00  \\
\hdashline
\multirow{2}{*}{TrTest} & GTr      & XLM-R & 27.43 & 29.53 & 29.76 & 26.42 & 28.29  & 10.61 & 19.55 & 18.70 & 14.94 & 15.95  \\
                                 & MarianMT   & XLM-R & 29.20 & 29.11 & 30.53 & 26.39 & 28.81               & 13.10 & 16.96 & 18.35 & 11.27 & 14.92   \\
\bottomrule
\end{tabular}
}%
}
\vspace{-0.5mm}
\caption{Per-language NLU results for two cross-lingual transfer methods: zero-shot cross-lingual transfer using multilingual pretrained models (MEncoder) and translate-test (TrTest) with Google Translate and MarianMT; see \S\ref{subsec: baselines} for more details. Translations for slot labelling were aligned using \textit{fast\_align} \cite{dyer2013fastalign}. The results of MEncoder are from the \textit{separate} training regime (see again \S\ref{subsec: baselines}). All scores are averages over 5 random seeds and follow the \textbf{All}-domain setup. Full results on the dev and test sets are provided in Appendix~\ref{app: detailed nlu}.}
\label{tab: different x-lingual transfers}
\vspace{-1mm}
\end{table*}

\subsection{Baselines and Experimental Setup}
\label{subsec: baselines}

We evaluate and compare the baselines for each task along the following axes: (i) different multilingual pretrained models; (ii) cross-lingual transfer approaches; (iii) in-domain versus cross-domain.

\rparagraph{Multilingual Pretrained Models} For cross-lingual transfer based on multilingual pretrained models, we abide by the standard procedure where the entire set of encoder parameters and the task-specific classifier head are fine-tuned. We evaluate the following pretrained language models: (i) for NLU and DST, we use the Base variants of multilingual BERT \citep[mBERT;][]{devlin2019bert} and XLM on RoBERTa \citep[XLM-R;][]{Conneau:2020acl}; for intent detection and slot labelling, we evaluate both a model that jointly learns the two tasks \cite{xu2020multiatis++} as well as separate task-specific models; (ii) for E2E modelling, we use multilingual T5 \citep[mT5;][]{mt5}, a sequence-to-sequence model, as it demonstrated to be the strongest baseline for cross-lingual dialogue generation \citep{lin2021bitod}.

\rparagraph{Cross-lingual Transfer} We focus on two standard methods of cross-lingual transfer: (i) transfer based on multilingual pretrained models and (ii) \textit{translate-test} \cite{hu2020xtreme}. In (i), a Transformer-based encoder is pretrained on multiple languages with a language modelling objective, yielding strong cross-lingual representations that enable zero-shot model transfer. In (ii), test data in a target language are translated into English via a translation system. To this end, we compare translations obtained via Google Translate (GTr)\footnote{\url{cloud.google.com/translate/docs/apis}} and MarianMT~\cite{mariannmt}. 

For end-to-end training, we set up two additional cross-lingual baselines, similar to~\citet{lin2021bitod}. In few-shot fine-tuning (FF), after the model is trained on the source language data (English), it is further fine-tuned on a small number of target language dialogues. In our experiments for FF, we use dialogues in the development set in each language as few-shot learning data. In mixed language pretraining~\citep[MLT;][]{lin2021bitod}, the model is fine-tuned on mixed language data where the slot values in the source language data are substituted with their target language counterparts. Unlike~\citet{lin2021bitod}, we do not assume the existence of a bilingual parallel knowledge base, which is unrealistic for low-resource languages. Hence, the translations of the slot values are obtained via MarianMT~\cite{mariannmt}.

\captionsetup[subfigure]{oneside,margin={-0.5cm,0.5cm},skip=-2pt}
\begin{figure*}[!t]
  \centering
      \begin{subfigure}[!ht]{0.33\linewidth}
        \centering
        \includegraphics[width=1.0\linewidth]{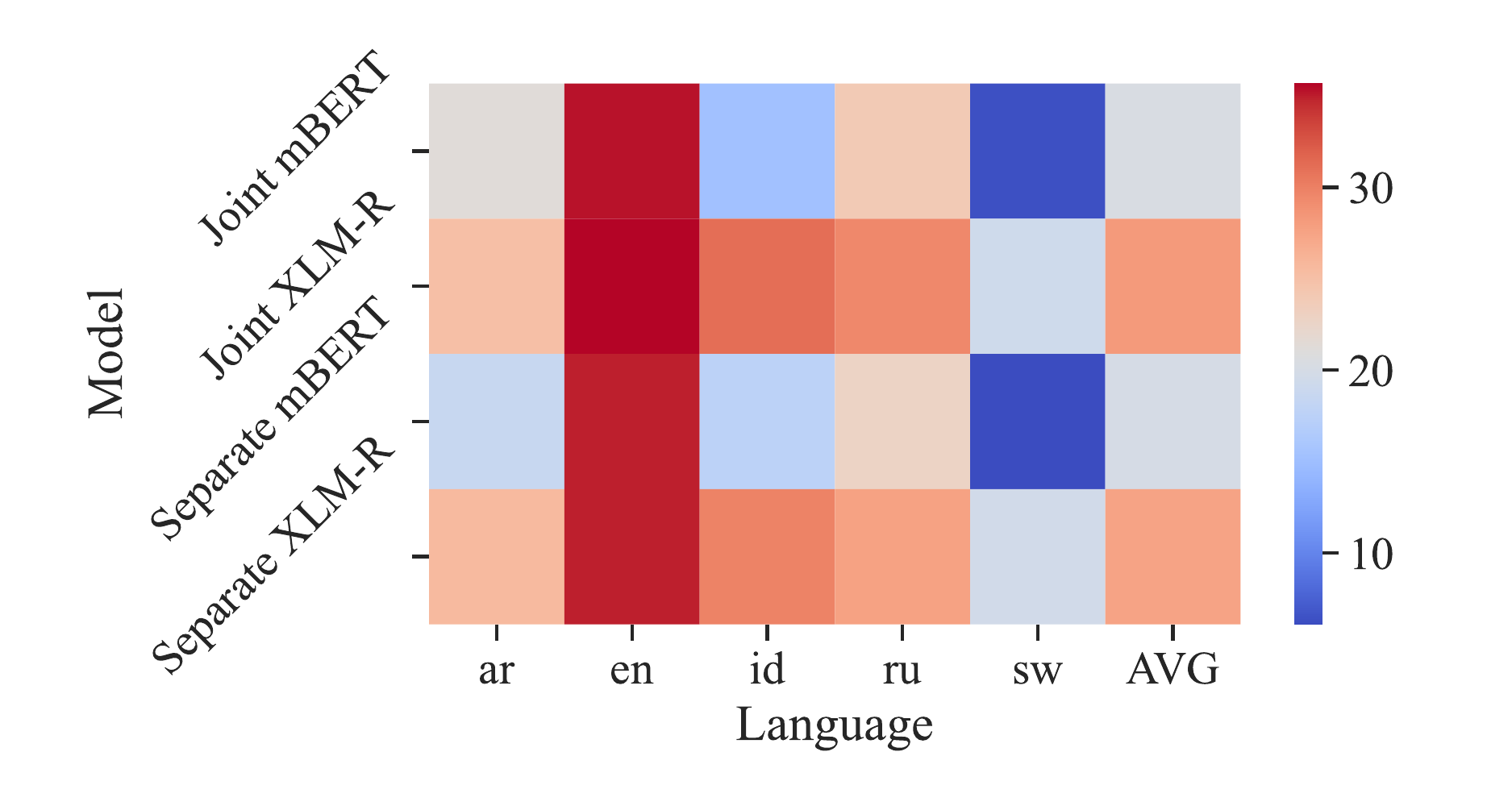}
        %\vspace{-0.7em}
        \caption{Intent Detection}
        \label{fig:intent_detection}
    \end{subfigure}
          \begin{subfigure}[!ht]{0.32\linewidth}
        \centering
        \includegraphics[width=1.0\linewidth]{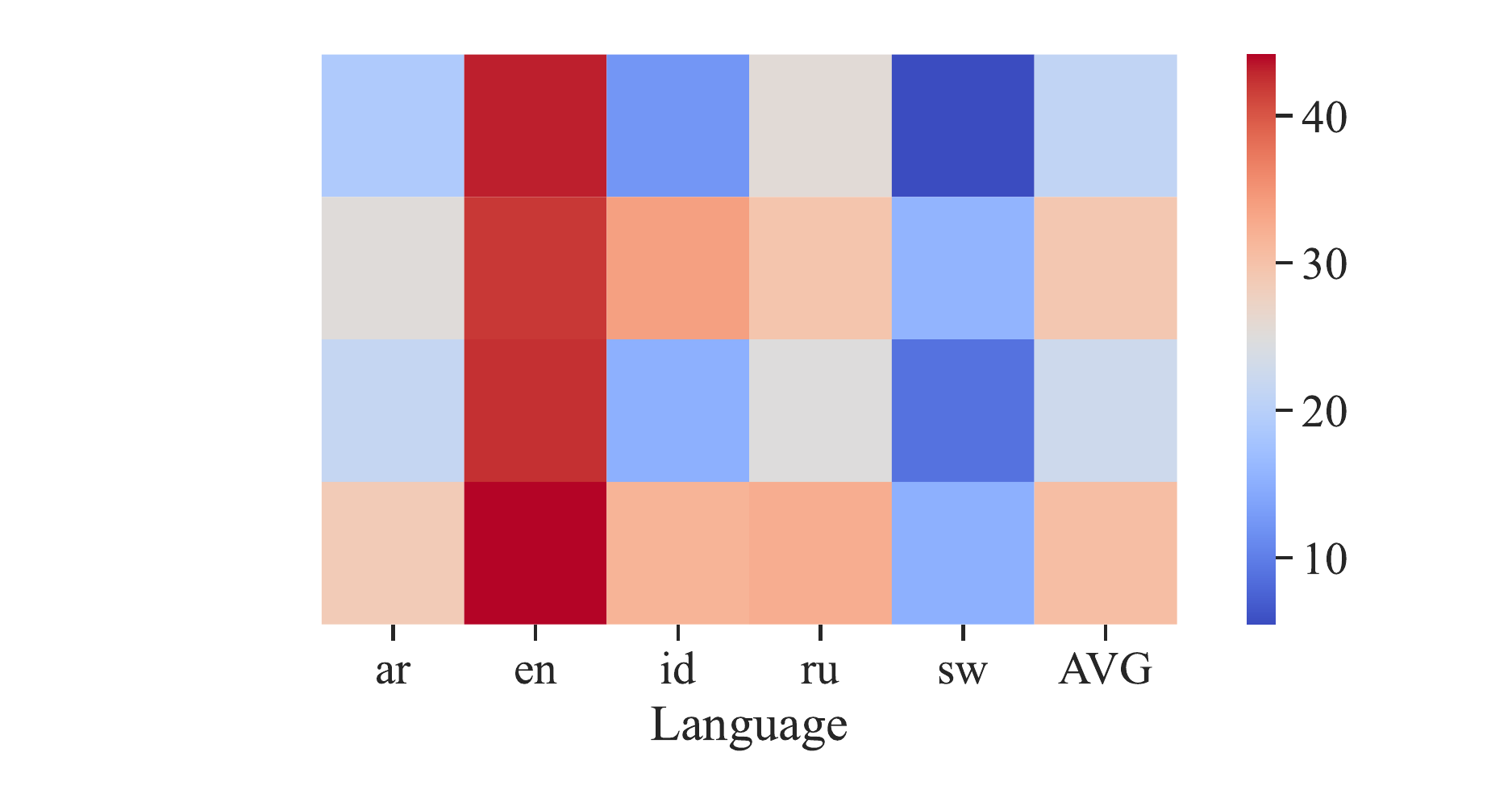}
        %\vspace{-0.7em}
        \caption{Slot Labelling}
        \label{fig:slot_heatmap}
    \end{subfigure}
              \begin{subfigure}[!ht]{0.33\linewidth}
        \centering
        \includegraphics[width=1.0\linewidth]{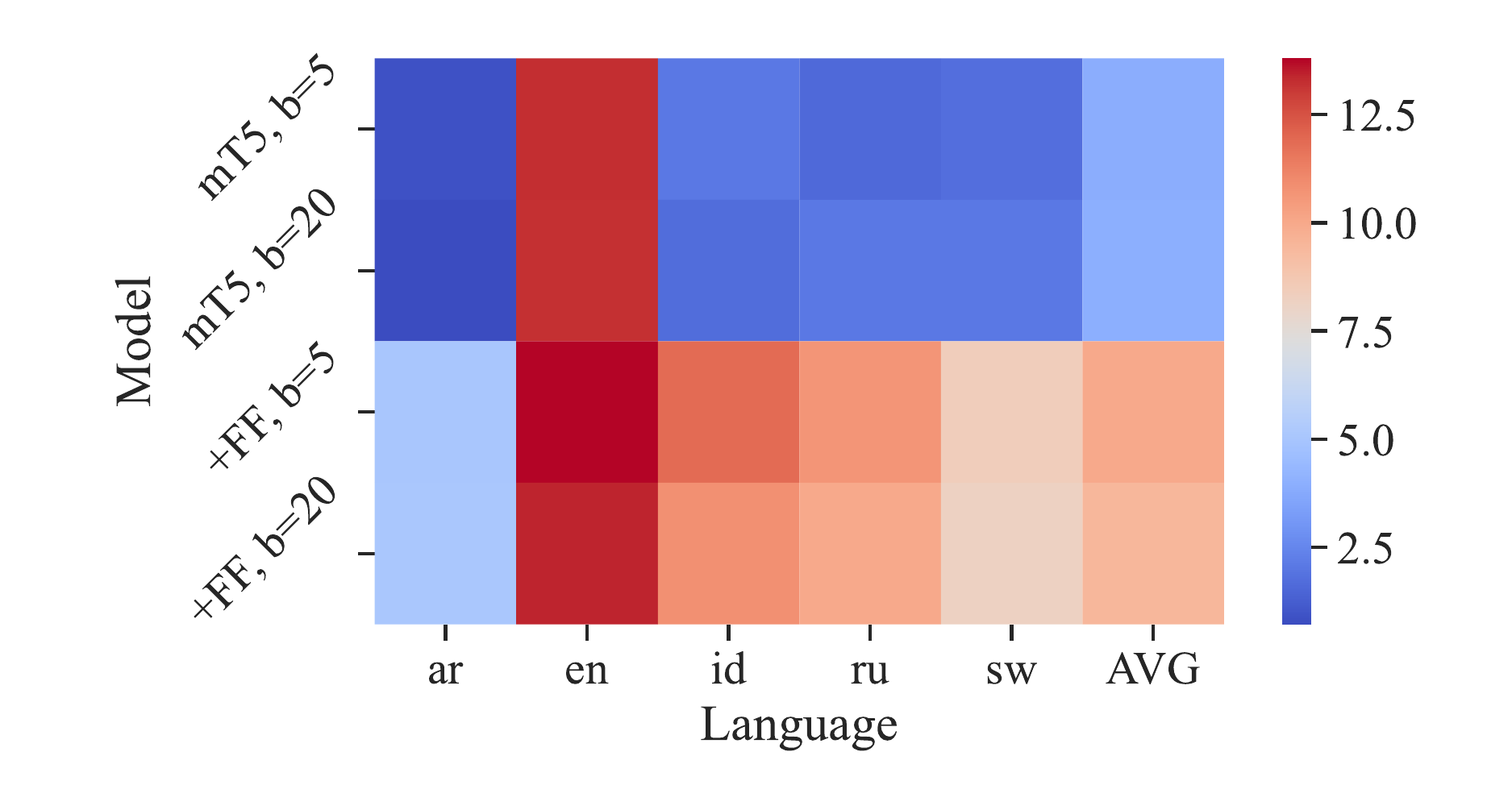}
        %\vspace{-0.7em}
        \caption{E2E}
        \label{fig:e2e_heatmap}
    \end{subfigure}
    \vspace{-0.5mm}
  %\subfigure[Intent Detection]{\includegraphics[width=0.35\textwidth]{Figures/intent_heatmap}}%
  %\subfigure[Slot Labelling]{\includegraphics[width=0.33\textwidth]{Figures/slot_heatmap}}%
  %\subfigure[E2E]{\includegraphics[width=0.35\textwidth]{Figures/e2e_heatmap}}
  \caption{Per-language results over all domains. (a) and (b) share the model labels on the y-axis.}
  \label{fig: per language nlu}
  \vspace{-1.5mm}
\end{figure*}

\begin{table}[!t]
\centering
\resizebox{\linewidth}{!}{%
{\footnotesize
\begin{tabular}{ll;{1pt/1pt}lllll} 
\toprule
\rowcolor{Gray}
\multicolumn{2}{l;{1pt/1pt}}{\begin{tabular}[c]{@{}l@{}}\\\end{tabular}} & \multicolumn{5}{c}{\bf E2E Training}                               \\
\rowcolor{Gray}
\textbf{Setup }              & \textbf{Model     }                          & AR   & ID   & RU   & SW   & AVG                                       \\ 
\cmidrule(l){1-7}
MEncoder                      & mT5                                 & 0.90 & 2.06 & 1.63 & 1.79 & 1.60                                      \\
\multicolumn{1}{r}{+FF}                        & mT5                                 & 4.36  & 10.96 & 8.48 & 7.79 &     7.90                                  \\
\multicolumn{1}{r}{+MLT +FF}                        & mT5                                 & 4.26 & 10.40 &  9.00 & 7.02 &  7.67                                     \\
\hdashline
TrTest \hspace{.5em} GTr      & mT5                                 & 1.87 & 1.96 & 4.38 & 2.59 & 2.59                                      \\
                        \multicolumn{1}{r}{MarianMT} & mT5                                 & 1.74 & 1.74 & 4.08 & 1.42 & \textcolor[rgb]{0.125,0.129,0.141}{2.25}  \\
\bottomrule
\end{tabular}
}%
}
\vspace{-1mm}
\caption{Per-language E2E results for two cross-lingual transfer methods (see also the information in Table~\ref{tab: different x-lingual transfers}).}
\label{tab:  x-lingual transfers e2e}
\vspace{-1.5mm}
\end{table}

\rparagraph{In-Domain versus Cross-Domain Experiments} \datasetacronym{} development and test splits include examples belonging to domains which were not seen in the English training data (see Table \ref{tab:domains}). This enables cross-lingual evaluation in 3 different regimes: \textit{in-domain} testing (\textbf{In}), where the model is evaluated on examples coming from the domains seen during training; \textit{cross-domain testing} (\textbf{Cross}), evaluating on examples coming from the domains which were \textit{not} seen during training; \textit{overall} testing (\textbf{All}), evaluating on all examples in the evaluation set. 

\rparagraph{Architectures and Training Hyperparameters} NLU in \tod consists of two tasks performed for each user turn $\mathcal{U}_i$: intent detection and slot labelling, which are typically framed as sentence- and token-level classification tasks, respectively. When a model is trained in a joint fashion, the two tasks share an encoder, and task specific classification layers are added on top of the encoder \cite{zhang2019jointlearningwithbert,xu2020multiatis++}. The loss is a sum of the intent classification and the slot labelling losses (cross-entropy). In \textit{separate} training, there is no parameter sharing, so neither NLU task influences the other. The performance metrics are \textit{accuracy} for intent detection and $F_1$ for slot labelling.

In the DST task, the model maps the dialogue history $\mathcal{H}_t$ to the belief state at $\mathcal{U}_{t}$; this includes the slot values that have been filled up to turn $t$. We use BERT-DST \citep{chao2019bert} in the experiments, which makes a binary classification regarding the relevance of every slot-value pair to the current context. During training, negative dialogue context-slot pairs are sampled randomly in a 1:1 ratio. At inference time, every context is mapped to every possible slot-value pair.

As in prior work \cite{lin2021bitod}, E2E modelling is framed as a seq-to-seq generation task. At every turn $t$, the goal is to predict the following $\mathcal{S}_t$ based on $\mathcal{H}_t$ fed into the model as a concatenated string. We adopt the generative seq2seq model, termed mSeq2Seq, as used by \citet{lin2021bitod}. This is based on mT5 Small \cite{mt5} and standard top-k sampling. As in prior work \cite{lin2021bitod}, performance is reported as BLEU scores \citep{papinenietal2002bleu}. Unless stated otherwise, we use a beam size of 5 for generation; see also Appendix~\ref{app:training hyperparameters} for further details.\footnote{We opt for mT5 as it substantially outperformed mBART \cite{liu2020multilingualBART} and other E2E baselines in the work of \citet{lin2021bitod}. We leave experimentation with more sophisticated model variants \cite{mintl} and sampling methods such as nucleus sampling \cite{nucleus_sampling} for future work. For brevity, we do not report results with other automatic metrics relevant to E2E modelling such as Task Success Rate or Dialogue Success Rate \cite{budzianowski2019hello}.}

\rparagraph{Source Language Training}
We train all models on the standard full training split of the English SGD dataset \cite{rastogi2020towards}. In order to measure performance gaps due to transfer and ensure comparability of dialogue flows in all languages, we also evaluate on the corresponding subset of the full English SGD test set, which was sampled as a source for the \cod dataset (see \S\ref{sec:design} and Table~\ref{tab:domains}).

%% (see Appendix~\ref{app:training hyperparameters} for other training hyper-parameters).

\subsection{Results and Discussion}
\label{subsec: results}

We now present and discuss the results of cross-lingual transfer under the experimental setups outlined in %Table \ref{tab: summary in cross domain} summarises the results on the test set. 
\textsection \ref{subsec: baselines}. We report both per-language scores and averages across the 4 \cod target languages.  %as it is the results on a corresponding subset of English dialogues rather than full dataset.} 

% included in \datasetacronym{},\footnote{We include the corresponding subset of \textsc{EN} SGD data.}

\rparagraph{Main Results}
Table~\ref{tab: different x-lingual transfers} compares the results for the two NLU tasks, while Table~\ref{tab:  x-lingual transfers e2e} shows the scores in the E2E task. Both include the two main methods of cross-lingual transfer, MEncoder and TrTest. With translate-test, the gains are highly task-dependent: it performs considerably better than encoder-based transfer methods on intent detection and E2E modelling, while the opposite holds for slot labelling. We speculate that this pattern stems from the following causes: \textbf{1)} we rely on a word alignment algorithm on top of English predictions to align them with the target language, which adds noise to the final predictions. \textbf{2)} Qualitative analysis of the predictions revealed that many errors are due to incorrect `label granularity’ (e.g., predicting \textit{departure city} instead of \textit{departure airport}).\footnote{This issue is more likely to occur in the case of translated text where language-specific hints for the exact slot type may get `lost in translation'.} Note that translate-test, unlike the encoder-based transfer transfer method, assumes access to high-quality MT systems and/or parallel data for different language pairs.

Table~\ref{tab:  x-lingual transfers e2e} reveals large gains of TrTest over the vanilla version of MEncoder. This occurs with both MarianMT and GTr, with GTr being the consistently better-performing translation method: this corroborates recent findings on other cross-lingual NLP tasks \cite{Ponti:2021translations}. However, the +FF results in Table~\ref{tab:  x-lingual transfers e2e} reverse this trend and underline the benefits of few shot target language fine-tuning in end-to-end training. The performance gains are large, even though the target language data includes only 92 dialogues (<1\% of English training data). In contrast, +MLT does not have a significant impact. This could be due to i) noisy target language substitutes, as they are obtained via automatic translation, unlike in \cite{lin2021bitod} where ground truth target language slot values were available; or ii) culture-specificity of slot values in \cod. Thus, substitution with translations appears to be beneficial only for dialogues with a pre-defined common cross-lingual slot ontology.

In DST, irrespective of the transfer method and target language, cross-lingual performance is near-zero (not shown). These findings are in line with prior work \citep{ding2021globalwoz} and are due to the DST task complexity. This is even more pronounced in zero-shot cross-lingual settings and especially for \cod, where culture-specific slot values are obtained via outline-based generation. Given the low results, we focus on NLU and E2E as the two main tasks in all the following analyses. 

%However, it is worth noting that overall the latter produces more consistent gains across languages and tasks.

\rparagraph{Comparison of Multilingual Models on NLU} The results in Table~\ref{tab: different x-lingual transfers} and Figure~\ref{fig: per language nlu} indicate that XLM-R largely outperforms mBERT in all setups in both NLU tasks. The gains are more pronounced on two languages more distant from English, \textsc{ID} and \textsc{SW}. We attribute this to XLM-R being exposed to more data in these languages during pretraining than mBERT. This very reason also accounts for the discrepancy in their performance on \textsc{EN} relative to other languages: with XLM-R, the gap between \textsc{EN} scores and other languages is much smaller than with mBERT. This is especially apparent in the case of Indonesian: \textsc{ID} pretraining data for mBERT is less than 10\% of \textsc{EN} pretraining data, while their sizes are comparable in XLM-R. 

Further, the results in Figure~\ref{fig: per language nlu} indicate that joint training of two NLU tasks tends to benefit intent detection while degrading the performance on slot labelling. The reverse trend is true for separate training: slot labelling scores improve, while intent detection degrades. This confirms the trend observed in recent work \cite{multisent}.\footnote{In another NLU experiment, we evaluated whether incorporating schemata into the NLU models---that is, leveraging short English descriptions of domains, intents, and slots available from the English SGD dataset---improves performance. We adapted the process of \citet{cao-zhang-2021-comparative} to a cross-lingual setup and obtained negative results, as using schemata yielded substantial performance drops. More details are provided in Appendix~\ref{app:schemata}.}

%\genie{Can we reference anonymous version of our paper with similar findings? Or is it not too double blind? :)  }

\rparagraph{Gaps with respect to English}
The per-language NLU results (see Table~\ref{tab: different x-lingual transfers} and Figure~\ref{fig: per language nlu}, and also Appendix~\ref{app: detailed nlu}) also illustrate the performance gap due to `information loss' during transfer: the drops (averaged across all 4 target languages) of the strongest transfer method are $\approx$10 points on intent detection (in \textbf{All}-domains experiments), and 15 points on slot labelling, using exactly the same underlying models. Moreover, the gaps are even more pronounced for some languages (e.g., Kiswahili as the lowest-resource language), and in domain-specific setups (e.g., in \textbf{In}-domain setups). 

In the E2E task, the results in Figure~\ref{fig:e2e_heatmap} also reveal a chasm between mT5 performance on English and the other four languages, especially so without any target-language adaptation. The gap, while still present, gets substantially reduced with the +FF model variant (see \S\ref{subsec: baselines}). This disparity emphasises the key importance of (i) continuous development of multilingual benchmarks inclusive of less-resourced languages to provide realistic estimates of performance on multilingual \tod, as well as (i) creation of (indispensable) in-domain data for few-shot target language adaptation. %The low scores indicate the complexity of the task in general.  

Overall, these findings suggest the challenging nature of the \cod dataset, and also call for further research on data-efficient and effective transfer methods in multilingual \tod. 

%\paragraph{Per-Language Results} In Figure \ref{fig: per language nlu} we report performance for each language. For NLU tasks, while the improved performance of XLM-R over mBERT is expected, 

%\paragraph{Cross-Lingual Transfer.} 

\begin{table}[!t]
\centering
\def\arraystretch{0.83}
{\footnotesize
\begin{tabularx}{\columnwidth}{ll YYY} 
\toprule
%\rowcolor{Gray}
\textbf{Method}                    & \textbf{Model} & \textbf{In}                        & \textbf{Cross }                    & \textbf{All  }                      \\ 
\midrule
\rowcolor{Gray}
\multicolumn{5}{c}{\textbf{Intent Detection} (\textit{Accuracy})}                                                                        \\ 
\midrule
\multirow{2}{*}{Joint}    & mBERT & 19.09                     & 16.92                     & 20.41                      \\
                          & XLM-R & {31.58} & {21.99} & {28.17}  \\
\multirow{2}{*}{Separate} & mBERT & {19.77} & {16.73} & {20.00}  \\
                          & XLM-R & 32.07                     & 20.94                     & 27.50                      \\ 
\midrule
\rowcolor{Gray}
\multicolumn{5}{c}{\textbf{Slot Labelling} ($F_1$)}                                                                                \\ 
\midrule
\multirow{2}{*}{Joint}    & mBERT & 23.46                     & 20.74                     & 21.13                      \\
                          & XLM-R & {53.35} & {26.56} & {37.45}  \\
\multirow{2}{*}{Separate} & mBERT & {25.72} & {21.93} & {22.58}  \\
                          & XLM-R & 53.36                     & 25.69                     & 36.75                      \\ 
\midrule
\rowcolor{Gray}
\multicolumn{5}{c}{\textbf{E2E} (\textit{BLEU})}                                                                                  \\ 
\midrule
                          & mT5   & 4.22                      & 3.76                      & 3.92                       \\
\bottomrule
\end{tabularx}
}%
\caption{Baseline results for NLU and E2E on the \datasetacronym{} test set, averaged over all 4 target languages, in three setups (\textbf{In}- or \textbf{Cross}-domain or \textbf{All} domains). Per-language results are in Appendix~\ref{app: detailed nlu}.}%\textbf{In:} in-domain testing. \textbf{Cross:} cross-domain testing. \textbf{All:} testing on all domains.}
\label{tab: summary in cross domain}
\end{table}

\rparagraph{In-Domain vs. Cross-Domain Evaluation.}
\label{para:crossdomain}
\cod not only enables cross-lingual transfer but is also the first multilingual dialogue dataset suitable for testing models in cross-domain settings: we summarise the results of this evaluation in Table \ref{tab: summary in cross domain}.
%The summary of results for in- and cross-domain settings are presented in Table \ref{tab: summary in cross domain}. 
%In the cross-domain setting, we train the model on the full training set and test it on the data from domains not represented in the training set. 
The key finding is that in-domain performance is much higher than cross-domain, although both have large room for improvement. %to our knowledge

\section{Related Work}
Although a number of NLU resources have recently emerged in languages other than English, the availability of high-quality, multi-domain data to support multilingual \tod is still inconsistent \cite{razumovskaia2021crossing}. Translation of English data has been the predominant method for generating examples in other languages. The ATIS corpus \citep{hemphill-etal-1990-atis} has been particularly widely translated,  boasting translations into Chinese \citep{he2013chineseatis}, Vietnamese \cite{dao2021vietnameseatis}, Spanish, German, Indonesian, or Turkish, among others \citep{susanto2017neural,upadhyay2018multiatis,xu2020multiatis++}. Bottom-up collection of \tod data directly in the target language has been the less popular choice, giving rise to monolingual datasets in French \citep{bonneau2005mediacorpus} and Chinese \citep{zhang2017first,gong2019escadataset}.

Thus far, the focus of existing benchmarks has been predominantly either on monolingual multi-domain \citep{hakkani2016multi,liu2019hwu64,Larson:2019emnlp} \textit{or} multilingual single-domain evaluation \citep{xu2020multiatis++}, rather than balancing diversity along both these dimensions. Moreover, the current multilingual datasets are mostly constrained to the two NLU tasks of intent detection and slot labelling \citep{li2020mtop,van-der-goot-etal-2021-masked}, and do not enable evaluations of E2E \tod systems in multilingual setups. In order to adequately assess the strengths and generalisability of NLU as well as DST and E2E models, they should be tested both on multiple languages \textit{and} multiple domains, a goal pursued in this work.

\section{Conclusion and Outlook}
In this work we have presented and validated a `bottom-up' method for creation of multilingual task-oriented dialogue (\tod) dataset. The key idea is to map domain-specific language-independent dialogue schemata into natural language outlines, which in turn guide human dialogue generators in each target language to create natural language utterances, both on the system and on the user side. We have empirically demonstrated that the proposed outline-based approach yields more natural and culturally more adapted dialogues than the standard translation-based approach to multilingual \tod data creation. Moreover, we have proven that the standard translation-based approaches often yield over-inflated and unrealistic performance in multilingual steps, while this issue is removed with the outline-based generation pipeline.

We have also presented a new \textbf{C}ross-lingual \textbf{O}utline-based \textbf{D}ialogue dataset (termed \datasetacronym), created via the proposed outline-based approach. The dataset covers 5 typologically diverse languages, 11 domains in total, and enables evaluations in standard NLU, DST, and end-to-end \tod tasks; this way, \cod makes an important step towards challenging multilingual \textit{and} multi-domain \tod evaluation in future research. We have also evaluated a series of state-of-the-art models for the different \tod tasks, setting baseline reference points, and revealing the challenging nature of the dataset with ample room for improvement. 

We hope that our work will inspire future research across multiple aspects. Besides its direct potential to serve as a more challenging testbed for current and future multilingual \tod models, our work provides useful practices and insights to steer and guide similar (potentially larger-scale) data creation efforts in \tod for other, especially lower-resource, languages and domains.

\section*{Acknowledgements}
{\scriptsize\euflag} This work has been funded by the ERC PoC Grant MultiConvAI: Enabling Multilingual Conversational AI (no. 957356) and a research donation from Huawei. The work of EMP is supported by the Facebook CIFAR AI Chair program.

\bibliography{anthology2021,dialogue}
\bibliographystyle{acl_natbib}

\clearpage
\appendix

%\label{sec:appendix}
\section{Dialogue Generation Guidelines}
\label{app:guidelines}
%\noindent\fbox{%
%    \parbox{\columnwidth}{%
Imagine having a conversation with a virtual or telephone assistant, where you want to complete a specific task. For example, you feel like going to a concert and would like to find out if there are any in your area, or would like to travel and need to book a flight.

In this task, we ask you to take on both roles, the user and the assistant: what would a helpful assistant reply to your query? Try and imagine an actual conversation you might have with an employee of a hotel or an airline, or at a tourist information office -- the aim is to write down natural conversations that could take place between two \textit{language\_name} speakers.\footnote{We provide a general guidelines template, where ``language\_name'' is a placeholder for the target language name.}

As a user, you will need to provide all the information that the assistant might need to carry out the task for you. You can be casual, like with someone you know and would address directly. As an assistant, you will provide information about flights, events, music, movies, or make suggestions that may interest the user.

In this task, we will provide you with brief instructions and types of information that the conversation between the user and the assistant should contain. However, to make the dialogues more natural to (hypothetical) \textit{language\_name} users, we encourage you to replace proper names which relate to English song titles, films, airline companies, cities, etc., with equivalents in \textit{language\_name}. You have complete freedom to make the replacements as you feel appropriate, as long as they are consistent within a single dialogue. See examples in Table \ref{tab:prompt_examples}.

It is likely that some concepts found in the English-language outlines do not exist in your culture or are unfamiliar to \textit{language\_name} speakers. Feel free to omit or creatively change these cases, so that the dialogues are fully understandable to \textit{language\_name} speakers.
 %   }%
%}

%% Removed as it is now merged into the main paper
\iffalse
\section{Multilingual Dialogue Datasets}
\label{app: multilingual datasets comparisons}
\begin{table}[h!]
\centering
\caption{Datasets used for comparison in linguistic diversity}
\begin{tabular}{ll} 
\toprule
Dataset & Reference     \\ 
\midrule
\multicolumn{2}{c}{NLU datasets}    \\ 
\cmidrule(lr){1-2}
  Multilingual TOP      & \citet{schuster2019cross}         \\
  Multilingual ATIS      &  \citet{upadhyay2018multiatis}         \\
  MultiATIS++     &      \citet{xu2020multiatis++}         \\
   MTOP     &      \citet{li2020mtop}         \\
    xSID    &        \citet{van-der-goot-etal-2021-masked}       \\ 
\midrule \midrule
\multicolumn{2}{c}{End-to-end datasets}    \\ 
\cmidrule(lr){1-2}
BiTOD        &      \citet{lin2021bitod}         \\
GlobalWOZ        &        \citet{ding2021globalwoz}       \\
\bottomrule
\end{tabular}
\end{table}
\fi

\newpage
\onecolumn
\section{Translation-Based versus Outline-Based Generation: Additional Examples}
\label{app: examples calques}
\begin{table}[!h]
\centering
\begin{adjustbox}{width=1\textwidth}
\begin{tabular}{lllll} 
\toprule
Dial. ID   & Example in \textbf{\textit{translation}}                                                                                & Example in \textbf{\textit{outline-generated}}                                                                                                & English example                                                                                                                                 & Type \& Comment on issue in \textbf{\textit{translation}}                                                                                                                                                                                                        \\ 
\midrule
1\_00030   & \begin{tabular}[c]{@{}l@{}} \foreignlanguage{russian}{Дата вылета и} \\\foreignlanguage{russian}{место назначения?}\end{tabular}                                               & \begin{tabular}[c]{@{}l@{}}\foreignlanguage{russian}{Конечно, на какое} \\\foreignlanguage{russian}{число и куда ты хотел} \\\foreignlanguage{russian}{бы полететь?}\end{tabular}                                             & Departure date and destination?                                                                                                                 & \vspace{-0.001 in} \multirow{2}{*}{\begin{tabular}[c]{@{}l@{}}$\spadesuit$: the word "\foreignlanguage{russian}{дата}" [date] is not used \\in spoken language; rather, "\foreignlanguage{russian}{какое} \\ \foreignlanguage{russian}{число}" [which number] is used\end{tabular}}                                                                                            \\
1\_00041   & \begin{tabular}[c]{@{}l@{}}\foreignlanguage{russian}{Откуда вылетаете и куда} \\\foreignlanguage{russian}{направляетесь? На какую дату} \\\foreignlanguage{russian}{вы хотели бы вылететь?}\end{tabular} & \begin{tabular}[c]{@{}l@{}}\foreignlanguage{russian}{Я могу помочь, откуда и куда}\\\foreignlanguage{russian}{полетите и на какое число} \\\foreignlanguage{russian}{мне искать билет?}\end{tabular}                          & \begin{tabular}[c]{@{}l@{}}Where would you like to leave \\from and where do you want to go? \\What date would you like to travel?\end{tabular} &        \\ [-2pt] \cmidrule(r){1-5}
1\_00090   & \begin{tabular}[c]{@{}l@{}}\foreignlanguage{russian}{Есть ли другие рейсы?} \\ \foreignlanguage{russian}{У меня 0 сумок для регистрации.}\end{tabular}                         & \begin{tabular}[c]{@{}l@{}}\foreignlanguage{russian}{Мне нужен билет без багажа.} \\ \foreignlanguage{russian}{Можем поискать еще другие} \\ \foreignlanguage{russian}{варианты?}\end{tabular}                                  & \begin{tabular}[c]{@{}l@{}}Hmm, are there any other flights? \\There are 0 bags for me to check in.\end{tabular}                                & \vspace{0.1 in} \multirow{2}{*}{\begin{tabular}[c]{@{}l@{}}$\spadesuit$: use of the number "0" is unnatural in \\spoken language; in outline-generated \\examples the speakers opt for \\"\foreignlanguage{russian}{без}" [without]. This seems like \\an artefact of translation being \\a text-to-text task.\end{tabular}}  \\
12\_00100  & \begin{tabular}[c]{@{}l@{}}Delta Airlines, \foreignlanguage{russian}{1 рейс в 9:15 утра, 207\$,} \\\foreignlanguage{russian}{0 пересадок~ ~}\end{tabular}                     & \begin{tabular}[c]{@{}l@{}}\foreignlanguage{russian}{Мне удалось найти один рейс} \\\foreignlanguage{russian}{Аэрофлотом без пересадки за} \\\foreignlanguage{russian}{15000 рублей. Отправление в}\\ \foreignlanguage{russian}{9:15 утра.}\end{tabular} & delta airlines 1 flight 9:15 am \$207 0 layovers                                                                                                &                                                                                                                                                                                                                                                                      \\ [-2pt] \cmidrule(r){1-5}
    3\_00113   & \begin{tabular}[c]{@{}l@{}}\foreignlanguage{russian}{Не хотите ли вы запланировать визит,} \\ \foreignlanguage{russian}{чтобы осмотреть недвижимость?}\end{tabular}            & \begin{tabular}[c]{@{}l@{}}\foreignlanguage{russian}{Вы хотели бы забронировать} \\ \foreignlanguage{russian}{осмотр квартиры на какую-нибудь дату?}\end{tabular}                                   & \begin{tabular}[c]{@{}l@{}}Do you want to schedule a visit to\\check out the property?\end{tabular}                                             &  \vspace{0.1 in} \multirow{2}{*}{\begin{tabular}[c]{@{}l@{}}$\spadesuit$: use of verb "\foreignlanguage{russian}{планировать}" [calque from \\ \textit{to plan}] instead of other more suited \\  options (e.g., "\foreignlanguage{russian}{хотели бы}" [would like to]) \end{tabular}}                                                                     \\
12\_00089~ & \foreignlanguage{russian}{На какой день планируете вылет?}                                                                                    & \foreignlanguage{russian}{В какой день вы бы хотели полететь?}                                                                                                          & Which is your preferred day of travel?                                                                                                          &                                                                                                                                                            \\ \cmidrule(r){1-5}
4\_00053   & \begin{tabular}[c]{@{}l@{}}\foreignlanguage{russian}{Меблирована ли квартира?} --\\\foreignlanguage{russian}{К сожалению, это не меблированная} \\\foreignlanguage{russian}{квартира.}\end{tabular}     & \begin{tabular}[c]{@{}l@{}}\foreignlanguage{russian}{Там есть мебель? --~Квартира без} \\ \foreignlanguage{russian}{мебели.}\end{tabular}                                                           & \begin{tabular}[c]{@{}l@{}}Is it furnished? -- Unfortunately, \\it isn't a furnished apartment.\end{tabular}                                    & \vspace{0.175 in} \multirow{2}{*}{\begin{tabular}[c]{@{}l@{}}$\clubsuit$: use of passive voice \\(e.g., "\foreignlanguage{russian}{меблирована}" [furnished]) \\which is rare in spoken language\end{tabular}}                                                                                                      \\
11\_00016~ & \foreignlanguage{russian}{Перевод начат!}                                                                                                       & \foreignlanguage{russian}{Трансакция прошла успешно.}                                                                                                                & Your transfer has been initiated!                                                                                                               &                                                                                                                                                                                                                                                                      \\ \cmidrule(r){1-5}
5\_00040   & \begin{tabular}[c]{@{}l@{}}\foreignlanguage{russian}{Можете ли вы найти мне} \\\foreignlanguage{russian}{музей в Лондоне, Великобритания?}\end{tabular}                       & \foreignlanguage{russian}{Хочу сегодня сходить в музей в Москве.}                                                                                                       & \begin{tabular}[c]{@{}l@{}}Can you find me a museum to \\visit in London, UK?\end{tabular}                                                      & \multirow{2}{*}{\begin{tabular}[c]{@{}l@{}}$\clubsuit$: use of a calque structure \\"\foreignlanguage{russian}{Можете ли вы найти мне}" \\{[}Can you find me]\end{tabular}}                                                                                                                                  \\
5\_00085~  & \begin{tabular}[c]{@{}l@{}}\foreignlanguage{russian}{Можете ли вы найти мне фильм,} \\ \foreignlanguage{russian}{режиссером которого является} \\ \foreignlanguage{russian}{Клер Дени?}\end{tabular}       & \foreignlanguage{russian}{Покажи фильмы с Максимом Матвеевым.}                                                                                                          & \begin{tabular}[c]{@{}l@{}}Can you find me a movie \\directed by Claire Denis?\end{tabular}                                                     &                                                                                                                                                                                                                                                                      \\
\bottomrule
\end{tabular}
\end{adjustbox}
\caption{Examples of unnatural linguistic choices in translations vs. outline-based generated sentences: $\spadesuit$ -- for choice of lexical cognates closer to source language; and $\clubsuit$ -- for syntactic calques from the source language. }
\end{table}

\twocolumn
\section{Additional Sentence Similarity Scores}
\label{app:cosine similarities encodings}
We also show additional (cosine) similarity scores between sentences generated via different data creation approaches (see \S\ref{ss:encodings} in the main paper) in Table~\ref{tab:more_cos} below.
\begin{table}[h!]
\centering
{\footnotesize
\begin{tabularx}{\columnwidth}{ll XX} 
\toprule
Split &  Encoder & \textbf{Translation }                                  & \textbf{Outline }                     \\ 
\cmidrule(lr){1-4}
Dev   & mDistilUSE\footnote{\citet{reimers-2019-sentence-bert};  Available at \url{https://huggingface.co/sentence-transformers/distiluse-base-multilingual-cased-v1} } & 0.80135    & 0.48491  \\
Test  & mDistilUSE & 0.78755 & 0.49853  \\
Dev   & LaBSE\footnote{\citet{feng2020language}; Available at \url{https://huggingface.co/sentence-transformers/LaBSE}} & 0.85850    & 0.54040  \\
Test  & LaBSE & 0.84416 & 0.55156  \\
Dev   & para-MiniLM\footnote{\citet{reimers-2019-sentence-bert}; Available at \url{https://huggingface.co/sentence-transformers/paraphrase-multilingual-MiniLM-L12-v2}.} &  0.83462   &  0.57392 \\
Test  & para-MiniLM & 0.84417 & 0.55156  \\
\bottomrule
\end{tabularx}
}%
\caption{Cosine similarities between encodings of English sentences, their translations to Russian (column Translation) and their counterparts generated based on outlines (Outline).}
\label{tab:more_cos}
\end{table}

\section{Training Hyper-Parameters}
\label{app:training hyperparameters}

\begin{table}[h!]
\centering
{\footnotesize
\begin{tabularx}{\linewidth}{l X} 
\toprule
\textbf{Parameter  }   & \textbf{Value }  \\ 
\midrule
Epochs        & 5       \\
Batch size    & 32  (8 for end-to-end)    \\
Learning Rate & 2e-5    \\
LR Scheduler  & Linear  \\
Weight Decay  & 0.01    \\
Optimizer     & Adam    \\
\bottomrule
\end{tabularx}
}%
\caption{Training hyper-parameters.}
\end{table}

\begin{table}[!t]
\centering
\caption{Results for schema-based intent prediction with mBERT based model.}
\label{tab: with schema intents}
{\footnotesize
\begin{tabular}{llllll} 
\toprule
\textbf{Schema?} & \textsc{AR}    & \textsc{ID}    & \textsc{RU}    & \textsc{SW}   & \textsc{AVG}                                        \\ 
\midrule
\textbf{Without}      & 18.61 & 17.57 & 22.83 & 6.09 & \textcolor[rgb]{0.125,0.129,0.141}{16.28}  \\
\textbf{With  }       & 14.50 & 8.88  & 14.20 & 4.14 & \textcolor[rgb]{0.125,0.129,0.141}{10.43}  \\
\bottomrule
\end{tabular}
}%
\end{table}

\section{Full NLU Results Per Language}
\label{app: detailed nlu}
Full (in-domain, cross-domain, all-domains) per-language results on \cod, with different NLU model variants based on multilingual encoders are provided in: 2) Table~\ref{intent:dev} (Intent classification, Development data); 1) Table~\ref{intent:test} (Intent classification, Test data); 3) Table~\ref{sl:dev} (Slot labelling, Development data);
4) Table~\ref{sl:test} (Slot labelling, Test data).

\begin{table*}
\begin{adjustbox}{width=1\textwidth}
\centering
\caption{Intent detection. Per-language results on the development set of the \cod dataset. The results are an average of 5 random seeds. \textit{in} corresponds to In-domain results; \textit{cross} corresponds to Cross-domain testing; \textit{all} denotes the results on All-domains. $\Delta$\textit{en} shows the gap between the results averaged across the four target languages (the AVG block) to the corresponding performance in English.}
\label{intent:dev}
%{\footnotesize
\begin{tabular}{llccc;{1pt/1pt}ccccccccccccccc;{1pt/1pt}ccc} 
\toprule
\multirow{2}{*}{\begin{tabular}[c]{@{}l@{}}\\\end{tabular}} & \multirow{2}{*}{\bf Model} & \multicolumn{3}{c}{en} & \multicolumn{3}{c}{ar} & \multicolumn{3}{c}{id} & \multicolumn{3}{c}{ru} & \multicolumn{3}{c}{swa} & \multicolumn{3}{c}{AVG} & \multicolumn{3}{c}{$\Delta$ en}  \\
                                                            &                        & in    & cross & all    & in    & cross & all    & in    & cross & all    & in    & cross & all    & in    & cross & all     & in    & cross & all     & in     & cross  & all                             \\ 

\cmidrule(lr){3-5} \cmidrule(lr){6-8} \cmidrule(lr){9-11} \cmidrule(lr){12-14} \cmidrule(lr){15-17} \cmidrule(lr){18-20} \cmidrule(lr){21-23}
\multirow{2}{*}{\bf Joint}                                      & mBERT                  & 55.83 & 40.45 & 51.56  & 28.49 & 26.07 & 29.32  & 33.03 & 30.94 & 31.71  & 31.29 & 22.25 & 28.57  & 14.16 & 7.48  & 9.88    & 26.74 & 21.69 & 24.87   & $-$29.09 & $-$18.77 & $-$26.69                          \\
                                                            & XLM-R                  & 54.02 & 40.90 & 50.93  & 46.29 & 30.19 & 41.97  & 46.82 & 35.96 & 44.57  & 38.26 & 32.66 & 39.33  & 36.06 & 29.74 & 33.71   & 41.85 & 32.14 & 39.90   & $-$12.17 & $-$8.76  & $-$11.03                          \\
\multirow{2}{*}{\bf Separate}                                   & mBERT                  & 53.48 & 40.08 & 50.06  & 31.14 & 26.97 & 28.93  & 35.38 & 28.54 & 32.02  & 35.98 & 20.15 & 30.70  & 17.00 & 7.50  & 11.95   & 29.87 & 20.79 & 25.90   & $-$23.61 & $-$19.29 & $-$24.16                          \\
                                                            & XLM-R                  & 55.08 & 41.05 & 51.49  & 46.82 & 30.41 & 42.50  & 45.76 & 35.66 & 43.69  & 39.17 & 33.41 & 40.25  & 32.80 & 30.34 & 32.72   & 41.14 & 32.46 & 39.79   & $-$13.94 & $-$8.59  & $-$11.70                          \\
\bottomrule
\end{tabular}
%}
\end{adjustbox}
\end{table*}

\begin{table*}
\centering
\caption{Intent detection. Per-language results on the test set of the \cod dataset. The results are an average of 5 random seeds. \textit{in} corresponds to In-domain results; \textit{cross} corresponds to Cross-domain testing; \textit{all} denotes the results on All-domains. $\Delta$\textit{en} shows the gap between the results averaged across the four target languages (the AVG block) to the corresponding performance in English.}
\label{intent:test}
\begin{adjustbox}{width=1\textwidth}
\begin{tabular}{llccc;{1pt/1pt}ccccccccccccccc;{1pt/1pt}ccc} 
\toprule
\multirow{2}{*}{\begin{tabular}[c]{@{}l@{}}\\\end{tabular}} & \multirow{2}{*}{\bf Model} & \multicolumn{3}{c}{en} & \multicolumn{3}{c}{ar} & \multicolumn{3}{c}{id} & \multicolumn{3}{c}{ru} & \multicolumn{3}{c}{swa} & \multicolumn{3}{c}{AVG} & \multicolumn{3}{c}{$\Delta$ en}  \\
                                                                            &                        & in    & cross & all    & in    & cross & all    & in    & cross & all    & in    & cross & all    & in    & cross & all     & in    & cross & all     & in     & cross  & all     \\ 
\cmidrule(lr){3-5} \cmidrule(lr){6-8} \cmidrule(lr){9-11} \cmidrule(lr){12-14} \cmidrule(lr){15-17} \cmidrule(lr){18-20} \cmidrule(lr){21-23}
\multirow{2}{*}{\bf Joint}                                                      & mBERT                  & 42.12 & 27.64 & 35.27  & 15.49 & 18.17 & 21.27  & 17.11 & 14.00 & 15.24  & 23.90 & 13.81 & 18.01  & 6.90  & 6.78  & 6.36    & 15.85 & 13.19 & 15.22   & $-$26.27 & $-$14.45 & $-$20.05  \\
                                                                            & XLM-R                  & 44.07 & 27.76 & 35.68  & 23.01 & 19.46 & 25.15  & 36.99 & 23.56 & 31.15  & 33.10 & 22.07 & 29.53  & 20.71 & 17.12 & 19.35   & 28.45 & 20.55 & 26.30   & $-$15.62 & $-$7.21  & $-$9.38   \\
\multirow{2}{*}{\bf Separate}                                                   & mBERT                  & 40.53 & 27.64 & 34.91  & 17.35 & 17.86 & 18.61  & 21.24 & 13.92 & 17.57  & 22.83 & 12.04 & 18.05  & 7.71  & 6.16  & 6.09    & 17.28 & 12.50 & 15.08   & $-$23.25 & $-$15.14 & $-$19.83  \\
                                                                            & XLM-R                  & 44.42 & 26.63 & 34.88  & 25.13 & 19.14 & 25.56  & 35.75 & 22.46 & 29.88  & 32.74 & 19.61 & 27.60  & 22.27 & 16.84 & 19.59   & 28.97 & 19.51 & 25.66   & $-$15.45 & $-$7.12  & $$-$$9.22   \\
\bottomrule
\end{tabular}
\end{adjustbox}
\end{table*}

\begin{table*}
\centering
\caption{Slot labelling. Per-language results on the development set of the \cod dataset. The results are an average of 5 random seeds. \textit{in} corresponds to In-domain results; \textit{cross} corresponds to Cross-domain testing; \textit{all} denotes the results on All-domains. $\Delta$\textit{en} shows the gap between the results averaged across the four target languages (the AVG block) to the corresponding performance in English.}
\label{sl:dev}
\begin{adjustbox}{width=1\textwidth}
\begin{tabular}{llccc;{1pt/1pt}ccccccccccccccc;{1pt/1pt}ccc} 
\toprule
\multirow{2}{*}{\begin{tabular}[c]{@{}l@{}}\\\end{tabular}} & \multirow{2}{*}{\bf Model} & \multicolumn{3}{c}{en}               & \multicolumn{3}{c}{ar} & \multicolumn{3}{c}{id} & \multicolumn{3}{c}{ru} & \multicolumn{3}{c}{swa} & \multicolumn{3}{c}{AVG} & \multicolumn{3}{c}{$\Delta$ en}  \\
                                                            &                        & in                   & cross & all   & in    & cross & all    & in    & cross & all    & in    & cross & all    & in    & cross & all     & in    & cross & all     & in     & cross  & all     \\ 
\cmidrule(lr){3-5} \cmidrule(lr){6-8} \cmidrule(lr){9-11} \cmidrule(lr){12-14} \cmidrule(lr){15-17} \cmidrule(lr){18-20} \cmidrule(lr){21-23}
\multirow{2}{*}{\bf Joint}                                      & mBERT                  & 80.5 & 38.36 & 55.22 & 34.77 & 21.86 & 26.62  & 22.72 & 16.16 & 18.47  & 50.06 & 16.49 & 28.65  & 23.45 & 6.46  & 12.88   & 32.75 & 15.24 & 21.66   & -47.79 & -23.12 & -33.56  \\
                                                            & XLM-R                  & 78.91                & 39.26 & 55.14 & 44.32 & 20.03 & 29.56  & 49.85 & 27.34 & 36.41  & 52.06 & 29.24 & 38.91  & 41.62 & 16.96 & 27.23   & 46.96 & 23.39 & 33.03   & -31.95 & -15.87 & -22.11  \\
\multirow{2}{*}{\bf Separate}                                   & mBERT                  & 81.27                & 34.12 & 53.07 & 39.54 & 22.83 & 29.09  & 22.82 & 18.00 & 19.66  & 45.79 & 17.08 & 29.07  & 21.27 & 6.61  & 12.27   & 32.36 & 16.13 & 22.52   & -48.91 & -17.99 & -30.55  \\
                                                            & XLM-R                  & 80.37                & 36.22 & 54.01 & 48.28 & 20.36 & 31.25  & 46.11 & 29.29 & 35.72  & 54.12 & 27.72 & 38.38  & 37.89 & 14.85 & 24.40   & 46.60 & 23.06 & 32.44   & -33.77 & -13.16 & -21.57  \\
\bottomrule
\end{tabular}
\end{adjustbox}
\end{table*}

\begin{table*}
\centering
\caption{Slot labelling. Per-language results on the test set of the \cod dataset. The results are an average of 5 random seeds. \textit{in} corresponds to In-domain results; \textit{cross} corresponds to Cross-domain testing; \textit{all} denotes the results on All-domains. $\Delta$\textit{en} shows the gap between the results averaged across the four target languages (the AVG block) to the corresponding performance in English.}
\label{sl:test}
\begin{adjustbox}{width=1\textwidth}
\begin{tabular}{llccc;{1pt/1pt}ccccccccccccccc;{1pt/1pt}ccc} 
\toprule
\multirow{2}{*}{\begin{tabular}[c]{@{}l@{}}\\\end{tabular}} & \multirow{2}{*}{\bf Model} & \multicolumn{3}{c}{en} & \multicolumn{3}{c}{ar} & \multicolumn{3}{c}{id} & \multicolumn{3}{c}{ru} & \multicolumn{3}{c}{swa} & \multicolumn{3}{c}{AVG} & \multicolumn{3}{c}{$\Delta$ en}  \\
                                                            &                        & in    & cross & all    & in    & cross & all    & in    & cross & all    & in    & cross & all    & in    & cross & all     & in    & cross & all     & in    & cross & all       \\ 
\cmidrule(lr){3-5} \cmidrule(lr){6-8} \cmidrule(lr){9-11} \cmidrule(lr){12-14} \cmidrule(lr){15-17} \cmidrule(lr){18-20} \cmidrule(lr){21-23}
\multirow{2}{*}{\bf Joint}                                      & mBERT                  & 42.12 & 42.89 & 43.23  & 23.64 & 17.54 & 19.02  & 20.27 & 11.19 & 12.40  & 20.28 & 24.21 & 25.52  & 11.00 & 7.89  & 5.47    & 18.80 & 15.21 & 15.60   & -23.32 & -27.68 & -27.63     \\
                                                            & XLM-R                  & 44.35 & 41.13 & 41.95  & 28.09 & 28.27 & 28.14  & 43.52 & 31.44 & 33.86  & 28.33 & 30.04 & 29.60  & 19.46 & 14.78 & 15.61   & 29.85 & 26.13 & 26.80   & -14.50 & -15.00 & -15.15     \\
\multirow{2}{*}{\bf Separate}                                   & mBERT                  & 41.62 & 42.74 & 42.36  & 26.35 & 20.36 & 21.54  & 24.20 & 12.72 & 15.30  & 20.85 & 26.08 & 24.89  & 15.60 & 7.77  & 8.84    & 21.75 & 16.73 & 17.64   & -19.87 & -26.01 & -24.72     \\
                                                            & XLM-R                  & 46.13 & 43.84 & 44.16  & 27.13 & 29.22 & 28.65  & 40.69 & 29.45 & 31.73  & 33.29 & 32.44 & 32.47  & 19.75 & 14.13 & 15.19   & 30.22 & 26.31 & 27.01   & -15.91 & -17.53 & -17.15     \\
\bottomrule
\end{tabular}
\end{adjustbox}
\end{table*}

\section{Leveraging SGD Schemata in NLU?}
\label{app:schemata}
Since the English SGD dataset \citep{shah2018building,rastogi2020towards} served as the starting point for \datasetacronym{}, we have access to its metadata (termed \textit{schemata}): short descriptions of domains, intents and slots released with SGD, provided in the English language. Leveraging such schemata was proven useful for boosting NLU results in monolingual English-only scenarios \cite{rastogi2020towards}. We thus evaluate whether incorporation of such schemata into the NLU models may positively impact their performance also in cross-lingual setups.    

For the intent detection task, we use domain and intent descriptions as the schema. Schemata are encoded with the multilingual pretrained model (mBERT) and are not fine-tuned in training, following the setup of \citet{rastogi2020towards}. To ensure comparability with results without schemata, we use only the user utterance as input into the intent classification model. At inference, we follow the process described by \citet{cao-zhang-2021-comparative}, where the schema for every intent is passed into the model together with the user utterance. The probability of the corresponding intent is recorded. If there is no intent with probability $>$0.5, \texttt{NONE} intent is predicted. This is a slightly different than our standard setup without the schema, where \texttt{NONE} intent is an additional class. 

The results in Table~\ref{tab: with schema intents} show that the use of schemata in cross-lingual settings does not provide performance boosts for intent prediction; on the contrary, we note a performance drop across the board. This could be a consequence of the increased number of trainable parameters due to the incorporation of schema embeddings into the model, which also might result in overfitting to the English training data.

\end{document}